\documentclass[lettersize,journal]{IEEEtran}
\usepackage{amsmath,amsfonts}
\usepackage{algorithmic, algorithm}
\usepackage{array}
\usepackage{amssymb}
\usepackage{subcaption}
\usepackage[caption=false,font=normalsize,labelfont=sf,textfont=sf]{subfig}
\usepackage{textcomp}
\usepackage{stfloats}
\usepackage{url}
\usepackage{verbatim}

\usepackage{graphicx}
\usepackage[capitalize,noabbrev]{cleveref}

\usepackage{xcolor}

\DeclareMathOperator*{\argmax}{arg\,max}
\DeclareMathOperator*{\argmin}{arg\,min}
\def\BibTeX{{\rm B\kern-.05em{\sc i\kern-.025em b}\kern-.08em
    T\kern-.1667em\lower.7ex\hbox{E}\kern-.125emX}}
\usepackage{balance}
\begin{document}
\title{Out-of-Distribution Adaptation in Offline RL: Counterfactual Reasoning via Causal Normalizing Flows}
\author{Minjae Cho, Jonathan P. How, and Chuangchuang Sun
\thanks{Minjae Cho is with the Department of Mechanical Engineering, Mississippi State University, MS 39762, USA (Email: \texttt{mc3216@msstate.edu}) \\
\indent Jonathan P. How is with the Department of Aeronautics and Astronautics, Massachusetts Institute of Technology, MA 02139, USA (Email: \texttt{jhow@mit.edu}) \\
\indent Chuangchuang Sun is with the Department of Aerospace Engineering, Mississippi State University, MS 39762, USA (Email: \texttt{csun@ae.msstate.edu})
}
}

\markboth{Journal of \LaTeX\ Class Files,~Vol.~18, No.~9, September~2020}%
{How to Use the IEEEtran \LaTeX \ Templates}

\maketitle

\begin{abstract}
Despite notable successes of Reinforcement Learning (RL), the prevalent use of an online learning paradigm prevents its widespread adoption, especially in hazardous or costly scenarios. Offline RL has emerged as an alternative solution, enabling learning from pre-collected static datasets, i.e., observational data. However, this offline paradigm introduces a new challenge known as ``distributional shift", degrading the performance when the policy is tested/ evaluated on scenarios that are Out-Of-Distribution (OOD) from the training dataset. Most existing offline RL approaches resolve this issue by regularizing policy learning within the information supported by the given dataset. However, such regularized learning overlooks the potential for high-reward regions that may exist beyond the dataset. This observation motivates exploring novel offline learning techniques that can generalize and make improvements beyond the data support without compromising policy performance, potentially by learning causation (cause-and-effect) instead of correlation from the dataset. In this paper, we propose the MOOD-CRL (Model-based Offline OOD-Adapting Causal RL) algorithm, which aims to address the challenge of extrapolation for offline policy training through causal inference instead of policy-regularizing methods. Specifically, Causal Normalizing Flow (CNF) is developed to learn the transition dynamics and reward function for data generation/ augmentation in offline policy evaluation and training. 
Based on the data-invariant, physics-based qualitative causal graph and the observational data, we develop a novel learning scheme for CNF to learn the quantitative structural causal model, leading to the understanding of the underlying dynamics. As a result, CNF is generalized with predictive and counterfactual reasoning capabilities for sequential decision-making tasks, revealing a high potential for OOD adaptation.
The effectiveness of our CNF-based offline RL approach is validated through extensive empirical evaluations, achieving comparable results with its online counterparts and outperforming both state-of-the-art model-free and model-based baselines with a significant margin.
\end{abstract}


\begin{IEEEkeywords}
Offline Reinforcement Learning, Causal Inference, Normalizing Flow, Counterfactual Reasoning
\end{IEEEkeywords}

\section{Introduction}
\label{submission}


\IEEEPARstart{W}{hile} Reinforcement Learning (RL) has achieved notable success in addressing sequential decision-making challenges across diverse domains, its deployment in real-world applications poses significant challenges due to the need for active/ online interaction with the environments during policy training. Specifically, online interactions can be costly, impractical, unethical, or prohibitive, especially in safety-critical domains like autonomous driving or healthcare. 
Meanwhile, in many applications, (large-scale) previously collected observational data is often available, such as power systems and autonomous driving.  
As a result, for an alternative solution to its online counterpart, offline RL has emerged to learn policies directly from a static observational dataset instead of online interaction. Yet, traditional offline learning encounters a significant obstacle known as \textit{distributional shift}. This shift occurs when the agent explores the part of state-action space that is barely included in the dataset. This results in extrapolation, where the values are typically overestimated. Consequently, the RL agent, aiming to maximize the return, may deteriorate in performance as it trusts in false belief.

Many efforts have been put to mitigate this distributional shift issue due to the increasing popularity of offline RL. One predominant way is to limit policy learning within the dataset by adopting extra conservatism in environment exploration \cite{rezaeifar2022offline, yu2020mopo}. This conservatism prevents the policy training and evaluation from straying too far from the data distribution to alleviate overestimation in uncertain regions. For more details, interested readers are referred to the \cref{sec:related_work} and the survey~\cite{levine2020offline}.

However, such a conservatism-based offline training scheme is likely to be sub-optimal, as the highest achievable performance is constrained by the given data. Hence, rather than confining the policy to learn solely within the data, a novel approach that is orthogonal to most existing works for offline RL is to facilitate exploration with accurate policy evaluation beyond the data support, i.e., out-of-distribution (OOD) adaptation. This is highly desirable when high-stake data points are impractical/ unethical to collect but at the same time are of high interest, such as vehicle crashes and certain clinical medical treatments. 
We argue that this can be achieved by utilizing available resources, including the dataset and other partially known/qualitative information, to infer the underlying environment dynamics and discover alternative optimal trajectories beyond the dataset. 

Our work aims to uncover regions with higher rewards without erroneous overestimation using causality-based approaches~\cite{pearl2009causality}, which avoids imposing any OOD regularizers and prevents overestimation by considering the cause-and-effect relationships among the variables of interest.
Specifically, our formulation excludes any elements that penalize exploration into OOD spaces, and the counterfactual reasoning at the highest level in Pearl’s causal hierarchy is employed for reasoning beyond the dataset. As a result, causal inference without OOD penalization \cite{pearl2010causal} brings the promise for exploration beyond the data support for OOD adaptation. Additionally, this type of Causal Reinforcement Learning (CRL), therefore, learns with causal relationships rather than relying solely on correlations within each provided dataset \cite{deng2023causal, zeng2023survey}. In the context of offline RL, instead of discovering causal relationships from data, we would like to learn the environment model, including the transition dynamics and reward, from the data assuming that a causal graph is given as a prior. Such a causal graph comes from the qualitative physics laws (e.g., force changes velocity based on Newton's second law) on top of the Markov decision process that formalizes sequential decision-making in RL.

To effectively learn a world model, i.e., the global data distribution beyond the given dataset, we combine causal inference and Normalizing Flows (NF), a generative model with exact likelihood quantification. Recent progress in NF offers substantial advantages due to its expressiveness and capacity for causal inference. Unlike Variational Autoencoders (VAE), normalizing flow integrates diffeomorphic transformations, which are both invertible and bijective. These properties enable transparent transformations, allowing OOD detection via probabilistic evaluation as well as molding simple distributions into more expressive multi-modal data distributions and vice versa. Moreover, studies in \cite{javaloy2023causal, khemakhem2021causal} present the equivalence of the autoregressive normalizing flow to acyclic causal models, showcasing its potential as a model for causal inference. Specifically, masking can eliminate dependency between irrelevant variables within each transformation layer of the autoregressive normalizing flow to preserve only interactions between relevant variables. Overall, leveraging the expressiveness, OOD detection, and causal inference capabilities of normalizing flow for dynamic modeling makes it an attractive choice for an under-explored regime of model-based offline RL with the promise to improve over observational datasets.

Building on these insights, we propose to learn the Markov Decision Process (MDP) by utilizing Causal Normalizing Flows (CNF) as a world model with provided causal graphs. In essence, facilitating OOD adaptation involves eliminating penalization for OOD instances while safeguarding against policy degradation resulting from OOD exploration. This is achieved by precisely identifying erroneous predictions that deviate from the dynamic nature of testing environments. Therefore, we introduce MOOD-CRL (Model-based Offline OOD-Adapting Causal RL) that leverages the aforementioned benefits to train policy. 

Our contributions are threefold as follows.
\begin{enumerate}
    \item We develop a model-based offline RL for OOD adaptation based on a causal model using normalizing flows. {This involves developing a novel learning scheme within the base distribution of NF, enabling counterfactual reasoning with a qualitative physics-informed causal graph, and augmenting data for offline policy training and evaluation.}

    \item We propose an effective model design using causal normalizing flow that systematically guides policy learning in a principled way. This includes leveraging OOD detection through accurate probabilistic density evaluation to prevent policy degradation caused by erroneous predictions exceeding the acceptable threshold.
    
    \item We perform comprehensive empirical investigations through various ablations, including: a) comparing with and without policy sophistication, b) evaluating against state-of-the-art model-based and model-free baselines, and c) testing across multiple robotic manipulation domains (Mujoco). This analysis provides valuable insights into our learning approach and demonstrates significantly enhanced performance compared to baselines, rivaling that of online methods.
    
\end{enumerate}

The remainder of this paper is organized as follows. The necessary background for our method is introduced in \cref{section:preliminaries}. Next, the details of our methodology and practical implementations are discussed in \cref{section:CNF}. This is followed by comprehensive experiments designed to evaluate the OOD adaptation of our approach, including extensive comparisons with other baselines, in \cref{section:experiments}. Finally, it is concluded with a discussion of the current limitations of our method and potential directions for promising future research to further advance offline RL in \cref{section:conclusions}.

\section{Preliminaries} \label{section:preliminaries}

\subsection{Reinforcement Learning}
Markov Decision Process (MDP) is a mathematical formulation for sequential decision-making problems. The MDP is a four-tuple $\langle \mathcal{S}, \mathcal{A}, T(s,a), r(s,a) \rangle$, including a state set $\mathcal{S}$, an action set $\mathcal{A}$, a transition function, $T(s,a): \mathcal{S} \times \mathcal{A} \rightarrow \mathcal{S'}$, and a reward function, $r(s,a): \mathcal{S} \times \mathcal{A} \rightarrow \mathbb{R}$. In reinforcement learning, the objective is to optimize the policy $\pi: \mathcal{S} \rightarrow \mathcal{A}$ by maximizing its cumulative reward $\sum_{t=0}^T \gamma^t r(s_t,a_t)$, where $\gamma\in [0,1]$ is the discount factor.

\subsection{Offline Reinforcement Learning}
In offline RL, the objective is identical to its online counterpart, but it only learns from a pre-collected dataset without online interaction, by either revealing the dynamic model (model-based) or directly finding a possibly optimal (often sub-optimal) policy $\pi$ (model-free). While this learning approach presents numerous advantages by decoupling its training reliance from constant environmental feedback, it introduces a new challenge: the policy's performance degrades when confronted with information not present in the dataset, a phenomenon referred to as \emph{'distributional shift'}. This factor is known to significantly impact the training process, as highlighted by \cite{levine2020offline}. Various works have explored this area from different angles, including the exploration of conservative value functions \cite{kumar2020conservative}, estimation of state-action density in policy training \cite{lee2021optidice}, and utilization of a structurally superior network architecture (specifically the Transformer \cite{chen2021decision}), yet no promising results compared to their online counterpart. Hence, a crucial yet under-explored aspect of offline (reinforcement) learning has been exploring approaches to address the distributional shift with generalizability and adaptation beyond static datasets. This can be achieved by learning a world model capable of extrapolation.

\subsection{Causal Reinforcement Learning}
Causal RL combines causal inference \cite{pearl2010causal} and RL, focusing on identifying {true physics-based interactions and dependencies among variables rather than statistical correlations.} In contrast, traditional RL assumes that all variables can interact with each other, which can lead to inaccurate analysis, known as spurious variables. For example, observing an increase in shark attacks at the shore alongside rising ice cream sales may suggest a correlation, but it does not necessarily imply causation without scientific evidence \cite{KVAMME2022119400}. Therefore, integrating causal analysis into the RL framework holds the promise of further enhancing RL in terms of both performance and explainability, as it has the potential to mitigate erroneous predictions when confronted with previously unseen scenarios.

Causal RL aims to address this issue by focusing on the interdependency between variables through causal discovery and inference~\cite{zhu2019causal, gasse2021causal, glymour2019review}. This process typically involves adopting a score metric and constructing a causal graph among variables to identify relationships with the highest score. Following this, the policy undergoes optimization to address spurious correlations and emphasize genuinely related ones. Additionally, another line of research explores learning based on the discovered causal relationships \cite{lu2022efficient}. Despite significant attention given to causal discovery, a gap exists in understanding how the discovered graphs can be effectively employed, particularly in offline learning. Consequently, this work assumes that causal graphs are given as priors. Usually, this causal graph is available following qualitative physics, such as that force changes velocity. As a result, this prerequisite is not restrictive in many cyber-physical systems, such as robotics and autonomous systems. In cases where a causal graph is unavailable, our method is applicable following causal discovery that learns a causal graph.

\subsection{Normalizing Flow}
Normalizing flow belongs to a distinct category of generative models, alongside Generative Adversarial Networks (GAN) and VAE. While sharing structural similarities with VAE, NF is different in incorporating diffeomorphic transformations and thus facilitating precise density evaluations. Unlike VAE's reliance on non-invertible neural networks, NF is bijective with invertible transformations that enable seamless passage of samples back and forth through the model, allowing probabilistic inference. Derived using the change of variable, the NF is updated by maximizing the model's probability density, represented as $p(\cdot)$, in generating samples from one simple distribution to another distribution as below.
\begin{equation}\label{eq:NF}
    \begin{aligned}
        p_x(\mathbf{x}) = p_u(\mathbf{u})|\text{det} J_F(\mathbf{u})|^{-1}, \quad \text{with } \mathbf{u} = F^{-1}(\mathbf{x}) \\
    \end{aligned}
\end{equation}
where $\mathbf{x} \subset \mathbb{R}^D$ and $\mathbf{u}\subset \mathbb{R}^D$ are vectors of the same dimensionality (bijective), with $\mathbf{x}$ representing the original dataspace and $\mathbf{u}$ representing the base distribution (i.e., uniform or multivariate Gaussian). The Jacobian $J_F$ is the $D \times D$ matrix of all partial derivatives of $F$, a diffeomorphic transformation regarding $\mathbf{u}$.

The role of the Jacobian, $J_F(\mathbf{u})$, is crucial as it shapes the original data into a base distribution. However, the high computational complexity of computing the determinant of the Jacobian, i.e, $\mathcal{O}(D^3)$, has led to a convention of employing transformations with a lower triangular matrix of Jacobian, significantly reducing the complexity to linear as $\mathcal{O}(D)$. This preference makes autoregressive structure inherently suitable and a natural choice for the flow model framework \cite{papamakarios2021normalizing}. 

Moreover, the autoregressive NF has undergone theoretical scrutiny as a framework for causal inference, exemplified by studies such as \cite{javaloy2023causal, khemakhem2021causal}. In this theoretical examination, the flow of data passage is constrained by existing causal relationships, which are represented by a binary adjacency matrix of the causal graph. This approach simply prevents the information sharing between irrelevant nodes within the transformation, $F(\mathbf{x})$, computing with only causally-related ones. Under certain conditions, this masking process guarantees the equivalence between autoregressive causal NF and its corresponding structural causal model.

\section{Causal Normalizing Flow for Dynamic Modelling} \label{section:CNF}
In this section, we present our conceptualization of CNF as a comprehensive world model for offline sequential decision-making, emphasizing its structural distinctions from conventional predictive models such as neural networks (Multi-Layer Perceptron: MLP). Specifically, our methodology can achieve OOD predictive capabilities within the generative network architecture of normalizing flows.

\begin{figure}[t]
    \centering
    \begin{minipage}{0.5\columnwidth}
        \centering
        \includegraphics[width=0.95\columnwidth]{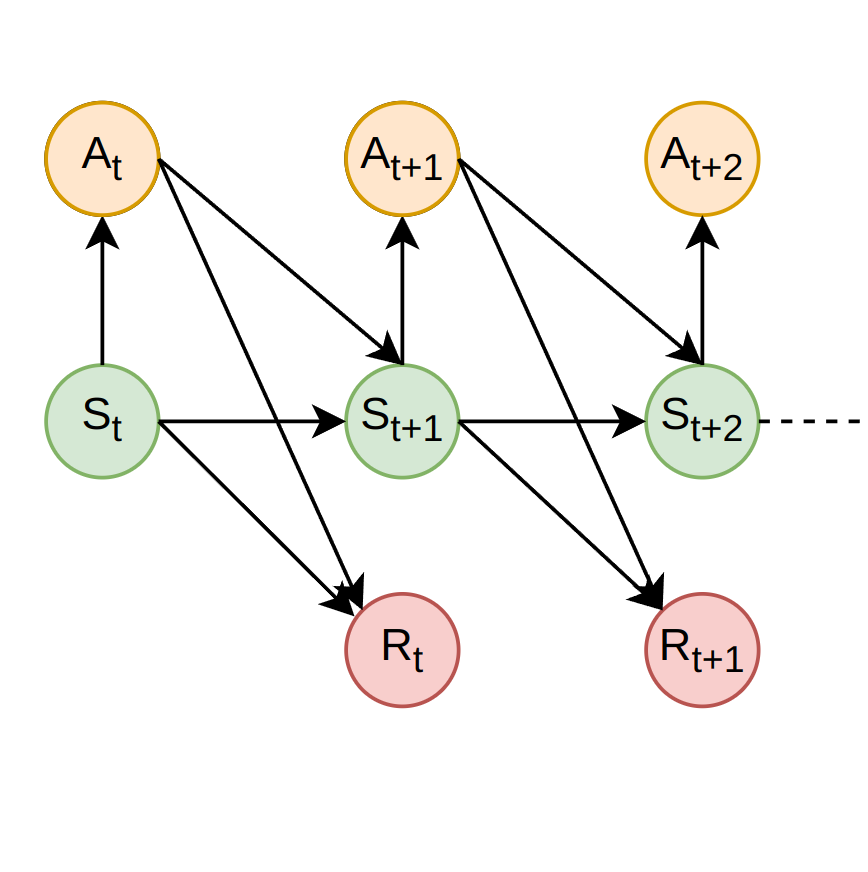}
        \subcaption{Causation in MDP}
        \label{fig:A}
    \end{minipage}%
    \begin{minipage}{0.5\columnwidth}
        \centering
        \includegraphics[width=0.95\columnwidth]{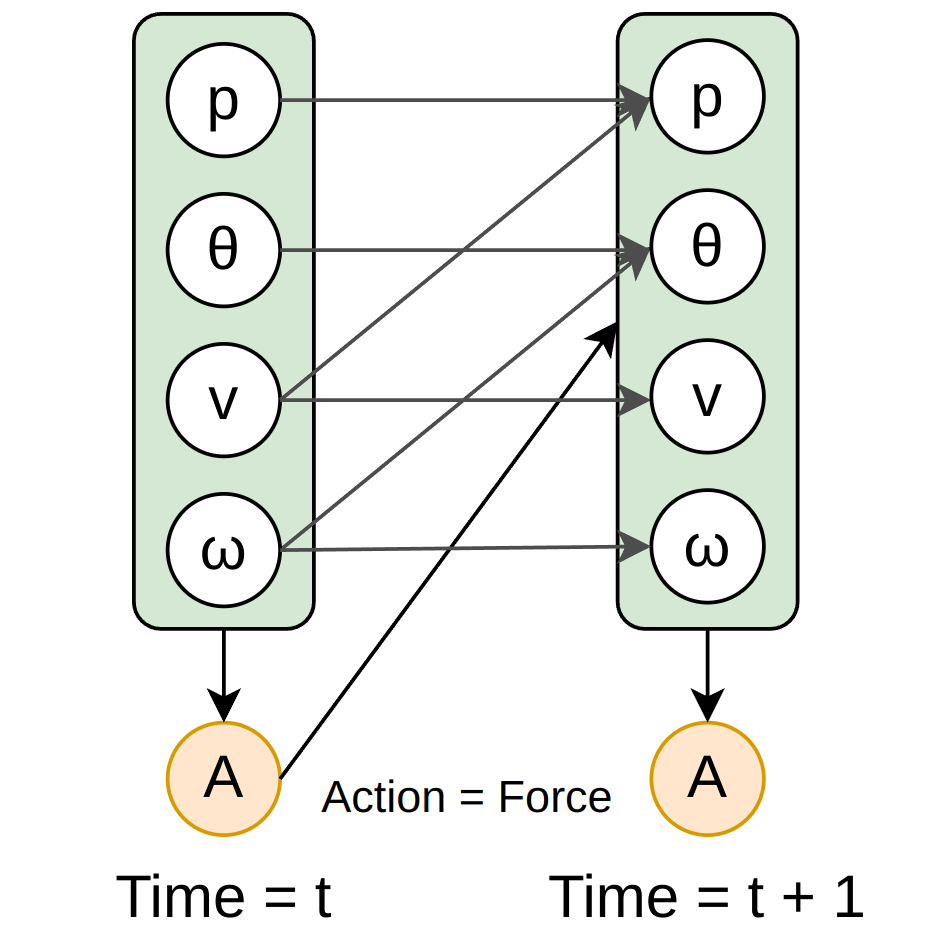}
        \subcaption{Causation in detail}
        \label{fig:B}
    \end{minipage}
    
    \caption{Illustration of causality in RL context. Figure 1(a) depicts the decision flow based on the Markov property: state causes action ($\mathcal{S} \rightarrow \mathcal{A}$), state-action causes next-state ($\mathcal{S} \times \mathcal{A} \rightarrow \mathcal{S}'$) and reward ($\mathcal{S} \times \mathcal{A} \rightarrow R$). \cref{fig:B} illustrates the qualitative physics-informed causation in state-to-state transitions for each time frame, specifically in an Inverted-Pendulum Environment by OpenAI Gym \cite{towers_gymnasium_2023}. The variables $p$, $\theta, v$, and $\omega$ correspond to position, angle, velocity, and angular velocity, respectively. As shown in \cref{fig:B}, the causation is grounded in physics, and multi-dimensional actions can be further partitioned into various parts for other environments.}
    \label{fig:envs}
\end{figure}

\subsection{Formulating CNF for MDP Prediction}
MLP can predict the next state and reward based on input states and actions, involving different dimensions of inputs and outputs (non-bijective), represented by $\mathbf{G}$. In contrast, the transformation within the CNF, denoted as $\mathbf{F}$, is characterized by a bijective process. Note that the bijective nature of CNF exposes limitations when applied as a world modeling tool in the context of RL. In standard scenarios, world models predict the next state and reward (dynamics) based on the previous state and action, which are mainly different dimensions of input and output. This insight reveals untapped potential in world modeling using normalizing flows, despite the clear advantages. Consequently, we introduce our formulation to facilitate world modeling, addressing its structural constraints.

The dataset in offline RL commonly comprises MDP tuples represented as $(s, a, s', r)$. To predict the dynamics of the environment using the bijective flow model, we utilize a bi-directional prediction process with two different MDP (true and perturbed) tuples with an additional mapping function to connect those in base space. This enables effective learning within the base distribution and harnesses the expressive power and OOD detection of normalizing flow. Specifically, we consider the true tuple $\mathbf{x} = (s, a, s', r)$ and the perturbed tuple $\tilde{\mathbf{x}} = (s, a, s, \varnothing)$, where the next-state and reward tokens are initialized as previous state, $s$, and zero (denoted as $\varnothing$), respectively. The rationale for this initialization approach is to align the base distributions of the perturbed tuple with the true one, thereby optimizing the mapping function for improved predictions. Essentially, by passing an initialized perturbed tuple through the flow model and employing a mapping function (in this case, an MLP), along with the inverse operation of the flow model, we aim to generate the expected true tuple.

These tuples, along with their corresponding base distributions $\mathbf{u}$ and $\tilde{\mathbf{u}}$, undergo the following sequential transformations:
\begin{equation}
    \tilde{\mathbf{x}} \xrightarrow{\mathbf{F}_{\text{CNF}}(\tilde{\mathbf{x}})} \tilde{\mathbf{u}} \xrightarrow{\mathbf{G}(\tilde{\mathbf{u}})} \mathbf{u} \xrightarrow{\mathbf{F}^{-1}_{\text{CNF}}(\mathbf{u})} \mathbf{x} 
\end{equation}

Here, $\mathbf{F}_{\text{CNF}}(\cdot)$ represents the transformation of the CNF, and $\mathbf{G}(\cdot)$ is a general function approximator mapping $\tilde{\mathbf{u}}$ to $\mathbf{u}$. They are composed as follows:
\begin{equation}
    \centering
    \begin{aligned}
        \mathbf{F}_{\text{CNF}} &= F_1 \circ \cdots \circ F_n \subset \mathbb{R}^{d \times d} \qquad \\
        \mathbf{G}_{\text{MLP}} &= G_1 \subset \mathbb{R}^{d \times m} \circ \cdots \circ G_n \subset \mathbb{R}^{n \times k},
    \end{aligned}
\end{equation} 
where $F_i,\forall i=1,\ldots n$ and $G_j,\forall j=1,\ldots,n$ are intermediate layers.

\begin{figure}[t!]
    \centering
        \includegraphics[width=1.0\linewidth]{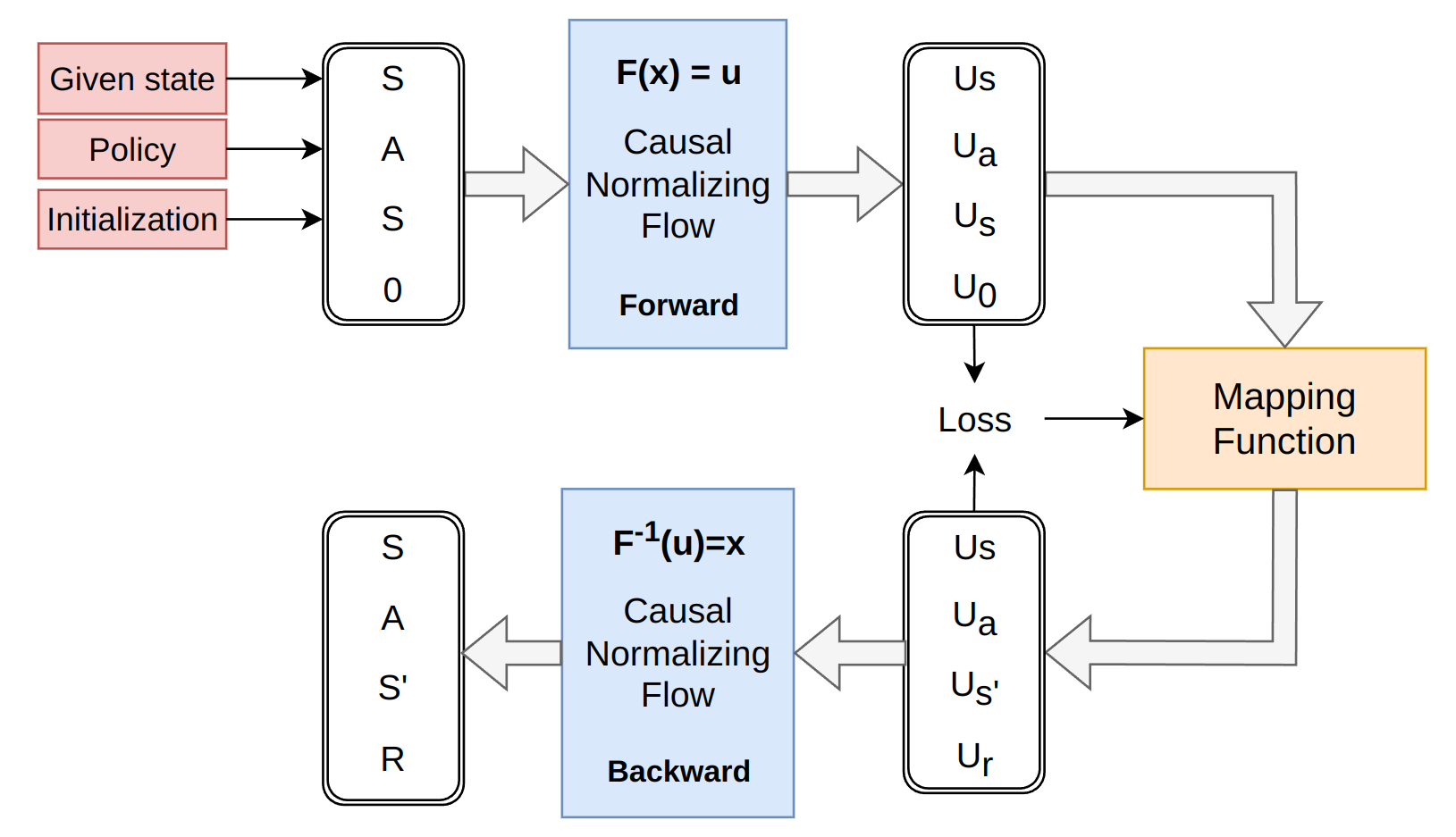}
        \caption{A depiction of Causal Reinforcement Learning employing a normalizing flow world model. The system handles two MDP tuples: the first is an arbitrary input for counterfactual reasoning, involving the use of state and action as inputs, initializing the next state with the current state for stability, and setting the reward to zero for initialization. The second tuple represents the original scenario. Subsequently, we calculate the loss between these tuples in the base space and train the mapping function.
        }
    \label{fig:CRL}
\end{figure}

\subsection{Architectural Details}
The diagram in \Cref{fig:CRL} illustrates our CNF framework for an offline model-based approach, depicting the dynamic interplay among the function approximator, MLP in the base space (serving as a mapping function to connect perturbed to true tuples), the flow model (deciphering predicted tuple in the base space to the true one), and the policy network (map from state to action for decision-making). Henceforth, we explain each component of our complete CNF model in full detail, following the aforementioned order with practical implementation and error analysis.

\subsubsection{Autoregressive Flow Model for Causal Inference}

As discussed in \cref{section:preliminaries}, the autoregressive framework is commonly embraced by flow models to capture dependencies among the variables and reduce computational costs. In our setup, we define $x$, $z$, and $u$ as the original data, intermediate data during the transformation process, and the ultimately transformed data in the base distribution, respectively. This process is articulated as follows: $F_1(\mathbf{x}) = \mathbf{z}_1, \cdots, F_k(\mathbf{z}_{k-1}) = \mathbf{z}_{k}, \cdots, F_n(\mathbf{z}_{n-1}) = \mathbf{u}$. Building on this, the autoregressive transformation, labeled $\mathbf{F}(\cdot|\tau, c^i)$, is defined by the subsequent expression:
$$z^i_{k+1}=\tau(z^{i}_k|\mathbf{h}^i), \text{ where } \mathbf{h}^i=c^i(\mathbf{z}_k^{<i-1})$$

\begin{figure}[t]
    \centering
    \includegraphics[width=0.95 \linewidth]{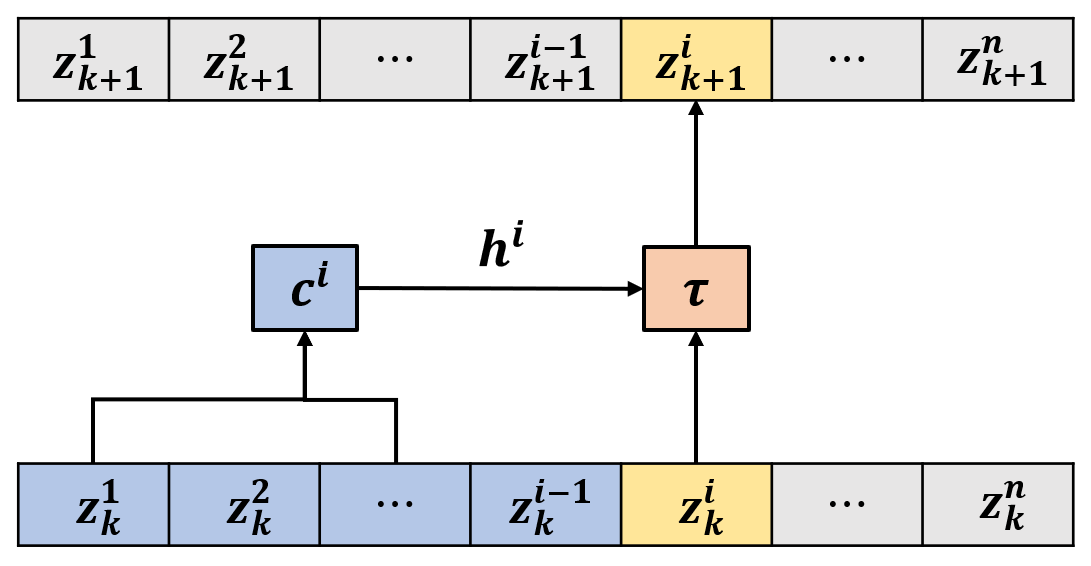}
    \caption{Autoregressive flow model illustration of transformation: $F_k(\mathbf{z}_k|\tau, c^i) = \mathbf{z}_{k+1}$. The conditioner, denoted as $c$, gathers information from previous elements, while $\tau$ transforms the current element and its history into a new value. The figure is adopted from 
    \cite{papamakarios2021normalizing}.}
    \label{fig:autoFlow}
\end{figure}

In this context, $\tau$ and $c$ denote the transformer and conditioner, respectively, as illustrated in \cref{fig:autoFlow}. The subscript, $k$, indicates the layer index, while the superscript, $i$, denotes the variable index within a layer. This framework inherently aligns with causal inference, particularly guided by an additional matrix-product operation inside $c$ (conditioner) with a causal graph. The schematic ensuring causation in the transformation is as follows:
\begin{equation}
    \begin{aligned}
        \mathbf{h}_i = c_i(\mathbf{z}^{<k}|A_{ij}^{<k})
    \end{aligned}
\end{equation}
In this context, the adjacency matrix denoted as $A_{ij}$ represents a causal graph specific to an RL problem, where $A \in \mathbb{R}^{d \times d}$ and $d$ is the size of the MDP tuple: $(s, a, s', r)$. This behaves as a masking of variables for those without relevance. It is worth noting that the adjacency matrix of the causal graph, a directed acyclic graph (DAG), can always be permuted to be a lower (or upper) triangle one. Taking the Inverted-Pendulum environment by OpenAI \cite{towers_gymnasium_2023} as an example, the states include the position of the cart, angle of the pole, velocity of the cart, and angular velocity of the pole. The action corresponds to the force applied to the cart to keep the pole upright. Leveraging the physics-based nature of the Inverted-Pendulum, an established causal graph between these variables is illustrated in \cref{fig:B}. 

The training of the flow model involves optimizing the parameters, $\theta$, of $\mathbf{F}(\cdot|\theta)$ to minimize the dissimilarity between the two distributions of true and generated samples via KL-divergence metric: $\text{KL}(p(\mathbf{x})||p_\theta(\mathbf{x})) = -\log p_\theta(\mathbf{x})$.
\begin{equation}
    \theta = \argmin \sum_{\mathbf{x} \in \mathcal{D}}  -\log p_\theta(\mathbf{x}).
    \label{eqn:cnf_loss}
\end{equation}
\subsubsection{Approximator in the Base Space}

Several methods have been investigated for manipulating the base space to achieve specific outcomes. One such example includes arithmetic operations to predict different angles of an image and applying constraints for improved classification \cite{radford2015unsupervised}. Following this appealing operation in base space, our approach is an explicit introduction of a mapping idea to enhance effective learning in the context of offline RL.

As previously mentioned, this involves using an MLP in the base space, which maps two MDP tuples, namely $\tilde{\mathbf{u}}$ and $\mathbf{u}$. The MLP is trained based on the discrepancy between these two tuples; for our case, we adopted the L1 norm as illustrated below:
\begin{equation}
    \phi = \argmin_\phi ||\mathbf{u} - G(\tilde{\mathbf{u}}|\phi)||_1 
    \label{eqn:mlp_loss}
\end{equation}
where the MLP model, parametrized by $\phi$, is trained using the provided dataset in the base distribution by passing it through the previously mentioned CNF.

\begin{algorithm}[t]
	\caption{MOOD-CRL: Model-based Offline OOD-Adapting Causal Reinforcement Learning}\label{alg:cap}
	\begin{algorithmic}[1]        
        \REQUIRE dataset $\mathcal{D}$, causal graph $A$
		\STATE \textbf{Initialization} Randomly initialize $\theta, \phi$
        \STATE \texttt{/*CNF update*/}		
    
        \FOR{each random batch in data, $B_i \in \mathcal{D}$}
            \STATE Update $\theta$ with \cref{eqn:cnf_loss} with Adam.
        \ENDFOR
        \STATE Create base data: $U = F(\mathcal{D})$ and $\tilde{U} = F(\mathcal{D}_{ptd})$
        
        \STATE \texttt{/*MLP update*/}
        
        \FOR{each random batch, $B_i \in U$, $\tilde{B}_i \in \tilde{U}$}
            \STATE Update $\phi$ with \cref{eqn:mlp_loss} with Adam.
        \ENDFOR
        \STATE \textbf{return} $F(\theta):= \text{CNF}(\theta)$ and $ G(\phi):=\text{MLP}(\phi)$
        
        \hrulefill        
        \texttt{/*Policy Training*/}\hrulefill 
        \REQUIRE $s, \pi$
        \WHILE{not done}
        \FOR{Each episode}
        \STATE Create Perturbed MDP $M_{ptd} = (s, \pi(a|s), s, \varnothing)$
        \STATE $M_{true}(s, a, s', r) \leftarrow F^{-1}\circ G \circ F(M) $
        \ENDFOR
        \STATE $\pi \leftarrow \argmax \mathbb{E}_{\substack{a_t \sim \pi,\\ (s',r) \sim \mathbf{\mathrm{CNF}}}} \sum_{t=1}^T r(s_t, a_t)$
        \ENDWHILE
	\end{algorithmic}
\end{algorithm}

\subsubsection{Policy Training with CNF}
Moving forward, our comprehensive training begins with initiating the training of the CNF model using the provided data, to employ it as a world model in lieu of any true model (e.g., OpenAI Gym environment). This marks a purposeful transition from the offline learning paradigm to an online approach, leveraging the pre-trained CNF as an oracle of transition dynamics and reward functions in the MDP. This implies that any online model-free method can be used as expressed below where transition and reward dynamics are estimated using CNF:
\begin{equation}
    \pi \leftarrow \argmax \mathbb{E}_{\substack{a_t \sim \pi,\\ (s',r) \sim \mathbf{\mathrm{CNF}}}} \Big[ \sum_{t=1}^T r(s_t, a_t) \Big]
\end{equation}
Crucially, unlike conventional model-based offline RL with an OOD penalty, our approach enables exploration into OOD spaces without incurring any OOD penalty. Instead, it effectively prevents policy degradation by accurately detecting erroneous predictions for previously unseen state-action pairs; see the implementation below.

\subsection{Implementation to Guide the Policy} \label{section:CNF_implementation}
In contrast to the previous offline RL approach \cite{kumar2020conservative,lee2022coptidice, yu2020mopo} where uncertainty estimation is used as a regularizer in reward, we do not include such penalization in our method. In Section \ref{section:experiments}, we demonstrate how this reward penalization deteriorates policy learning, resulting in inconsistent performance across diverse tasks. Moreover, to guarantee that our causal model makes reasonable predictions, we devise a principled approach to prevent deterioration in policy exploration, evaluation, and learning due to erroneous predictions in OOD space. This is achieved by terminating the episodic task when a certain OOD measure exceeds a reasonable threshold, benefiting from the power of uncertainty quantification of the CNF model. In other words, we filter out high-error/ low-probability predictions during policy learning to facilitate effective learning, leveraging exact probabilistic evaluation of normalizing flow. The terminating criterion is shown below:
\begin{equation} \label{eqn:truncate}
    \text{truncate}= 
\begin{cases}
    \text{True}, & \text{if } \log p_\theta(s_t,a_t, s_{t+1}, r_t) < c\\
    \text{False} ,              & \text{otherwise}
\end{cases}
\end{equation}

Here, the parameter $c$ is user-defined to control the level of exploration into OOD scenarios, and it was chosen as a negative integer. In \cref{alg:cap}, we provide pseudo-code for our algorithm termed MOOD-CRL, to represent the entire framework.

\section{Experiments} \label{section:experiments}
\begin{figure}
    \centering
    \includegraphics[width=0.6\linewidth]{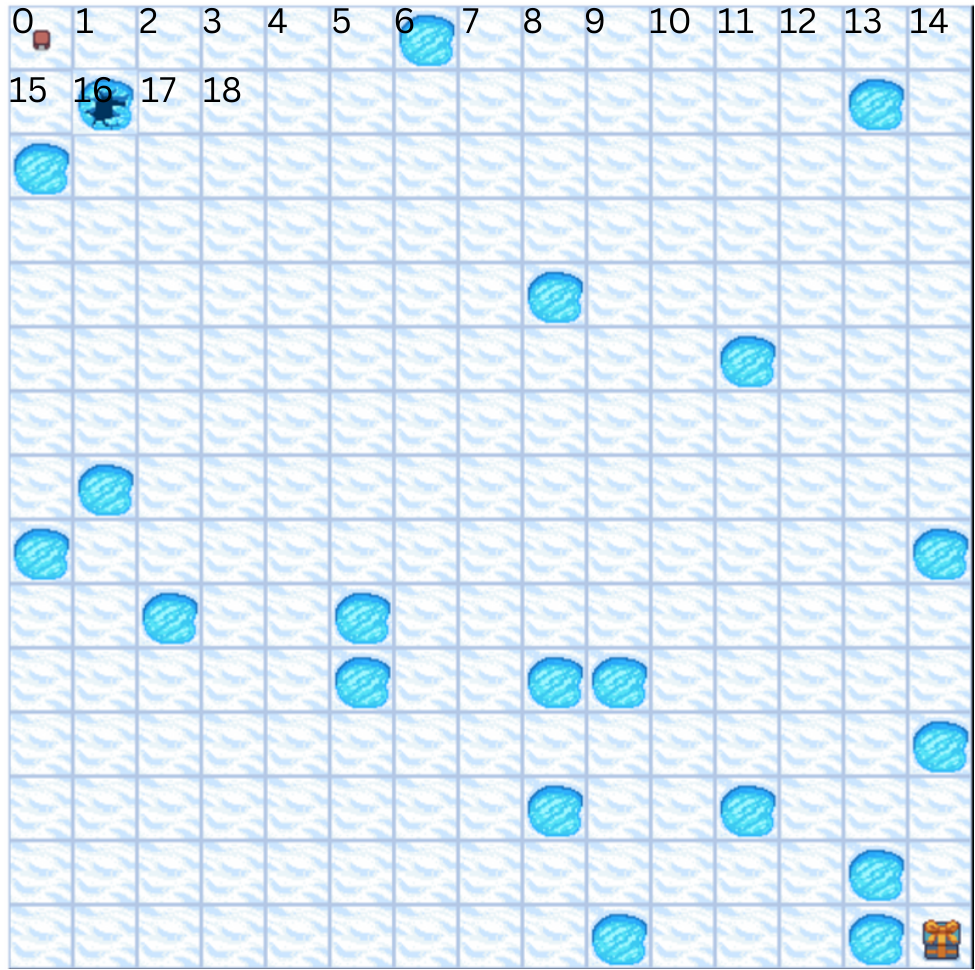}
    \caption{A classical toy problem implemented in OpenAI Gymnasium involves a 15 $\times$ 15 grid with non-slippery conditions selected for a deterministic environment. The top-left corner has a state of 0, increasing by 1 towards the right. Any actions towards boundaries and obstacles result in staying in the same position. }
    \label{fig:frozenlake}
\end{figure}

This section systematically and extensively evaluates our method using OpenAI's Gymnasium environments~\cite{towers_gymnasium_2023} to address the following questions.
\begin{enumerate}
    \item Is MOOD-CRL able to effectively capture causation among variables?
    \item Is MOOD-CRL able to surpass previous offline RL methodologies in achieving higher returns?
    \item Is MOOD-CRL able to operate under different qualities of data and different RL algorithms?
\end{enumerate}

\subsection*{Baselines Algorithms}
For extensive analysis, our baselines include comparisons with both model-based and model-free approaches, specifically the non-causal version of our approach (MOOD-RL), the state-of-the-art Model-based Offline Policy Optimization: MOPO \cite{yu2020mopo}, the state-of-the-art model-free policy optimization: OptiDICE \cite{lee2021optidice}, and traditional MLP-based predictions (for transition dynamics and rewards). The details are presented below:

\begin{itemize}
    \item \textbf{Model-based Offline OOD-Adapting RL (MOOD-RL)}: MOOD algorithm without a causal graph. It is simply base distributional learning in normalizing flows parameterized by MLP.
    \item \textbf{Model-based Offline Policy Optimization (MOPO)} \cite{yu2020mopo}: MLP-based network architecture which learns world dynamics and uncertainty present in predictions to avoid OOD exploration by reward penalty.
    \item \textbf{Offline Policy Optimization via Stationary Distribution Correction Estimation (OptiDICE)} \cite{lee2021optidice}: Model-free algorithm that directly learns a policy without learning transition dynamics and reward function via stationary distribution correction estimation. 
    \item \textbf{Multi-Layer Perceptrons (MLP)}: Traditional deep neural network architecture with non-linearity to predict the transition dynamics and reward.
\end{itemize}

\begin{figure}[t!]
    \centering
     \includegraphics[width=1\linewidth]{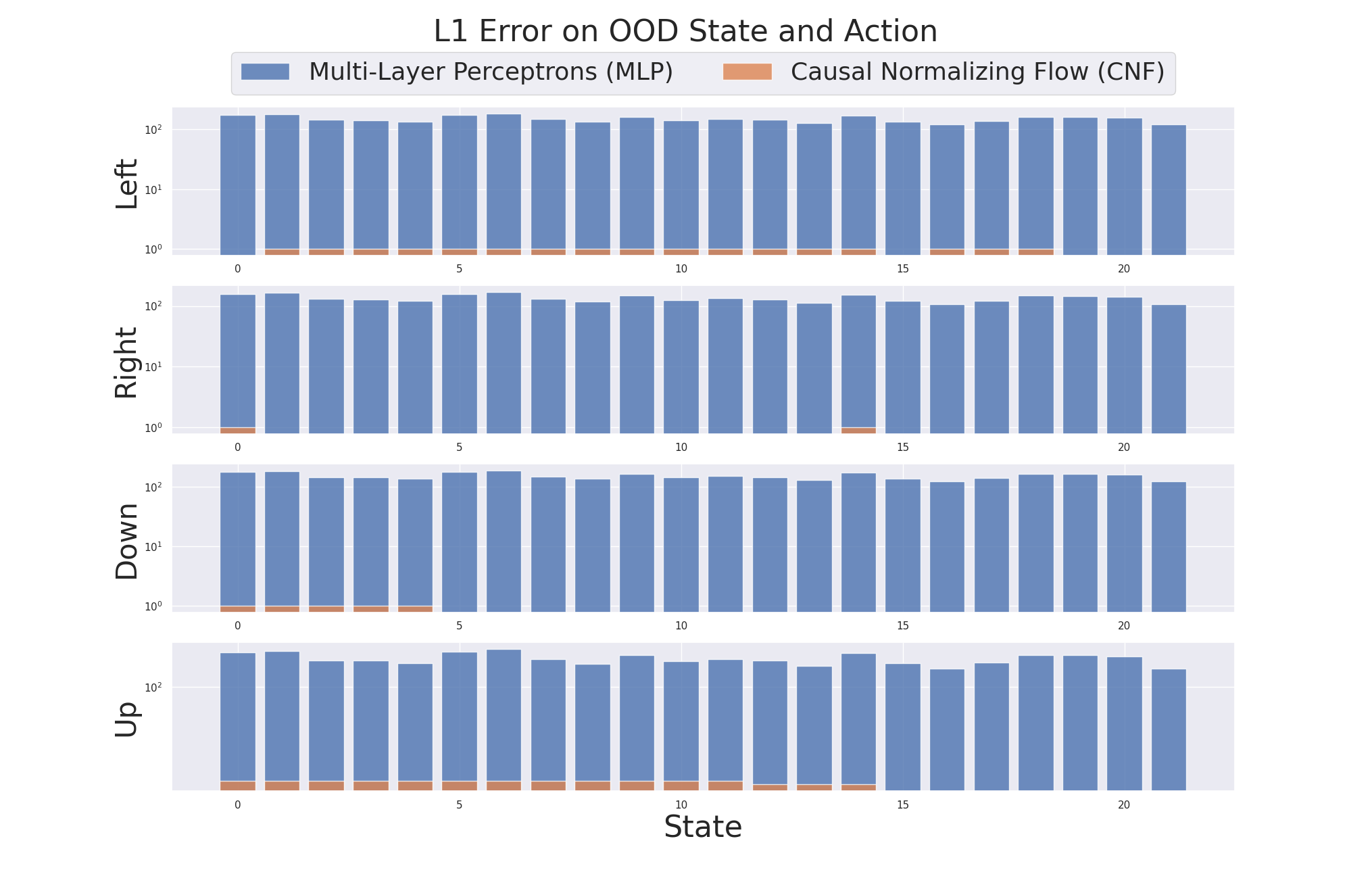} 
     \caption{A comparison of estimation errors between MLP estimator and MOOD-CRL (in log-scale). The training dataset consists of state-action combinations from $45 \leq s \leq 224$ with $0 \le s\le 44$ as the testing dataset. The results reveal that MOOD-CRL consistently delivers reasonable predictions in OOD scenarios. However, MOOD-CRL makes an error only for the 'Up' action at the top row of the grid, where inference is inherently challenging. In contrast, the MLP exhibits failures when confronted with OOD situations, yielding erroneous values.}
    \label{fig:Discrete}
\end{figure}

\subsection*{Testing Environments}
Moreover, we introduce our continuous environments here to answer questions 2 and 3. These environments are commonly used in RL, online or offline.
\begin{itemize}
    \item \textbf{Inverted Pendulum: } $\mathcal{S} \subset \mathbb{R}^4$ and $\mathcal{A} \subset \mathbb{R}^1$ where a cart with the vertical pole seeks to maintain the pole in an upright position and prevent it from falling.
    \item \textbf{Hopper: } $\mathcal{S} \subset \mathbb{R}^{11}$ and $\mathcal{A} \subset \mathbb{R}^3$ where a one-legged agent trying to achieve forward velocity without falling.
    \item \textbf{Walker: } $\mathcal{S} \subset \mathbb{R}^{17}$ and $\mathcal{A} \subset \mathbb{R}^6$ where a two-legged agent trying to achieve forward velocity without falling.    
    \item \textbf{HalfCheetah: } $\mathcal{S} \subset \mathbb{R}^{17}$ and $\mathcal{A} \subset \mathbb{R}^6$ where a two-legged cheetah learns to run forward without falling.
\end{itemize}

\begin{figure*}[t]
    \centering
    \begin{tabular}{c c c c c}
        \hline\hline
        \multicolumn{4}{c}{Return distribution }\\
        \includegraphics[width=0.225\linewidth]{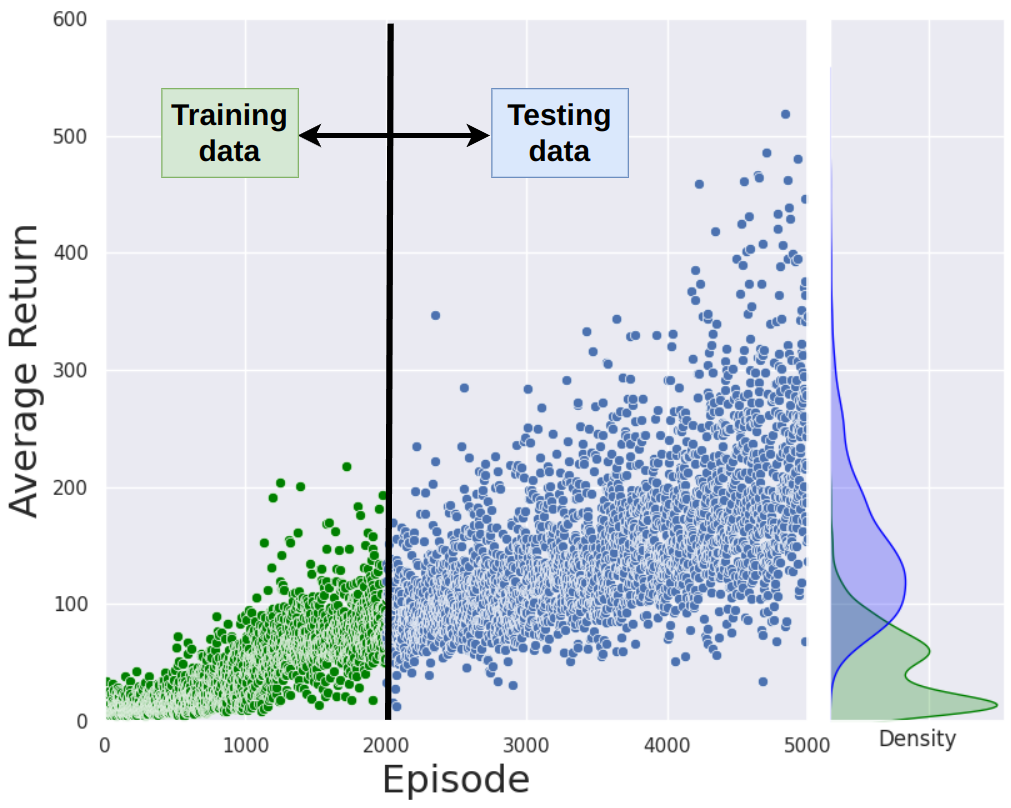}    &
        \includegraphics[width=0.225\linewidth]{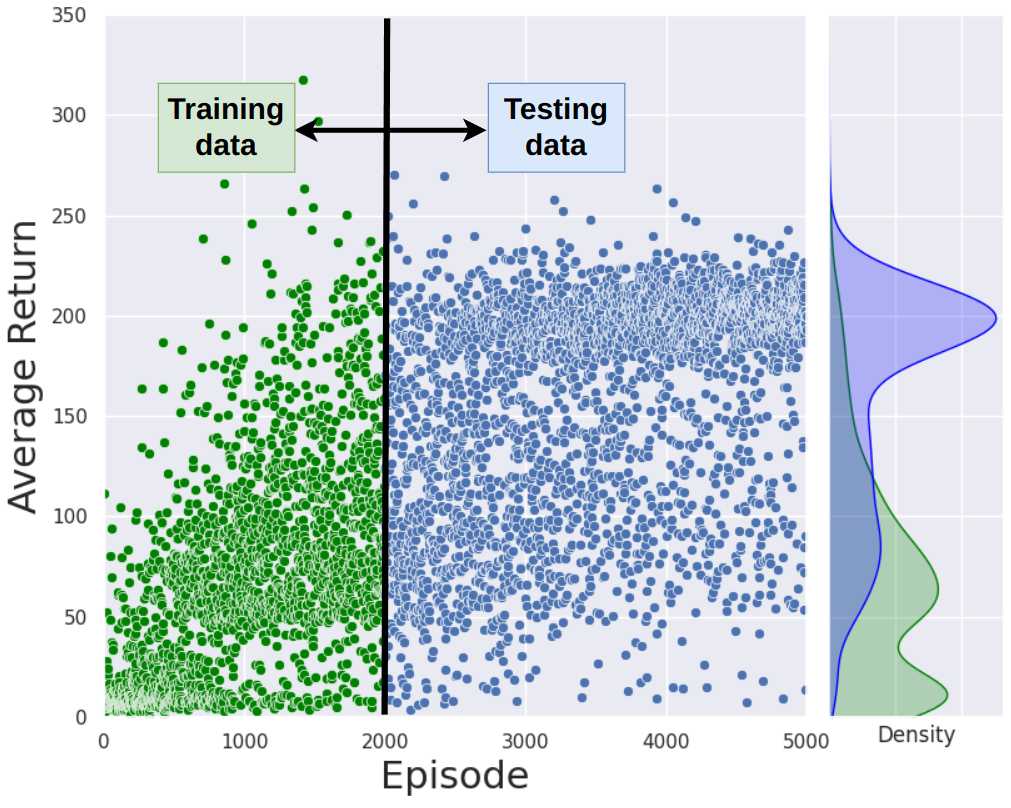}   &
        \includegraphics[width=0.225\linewidth]{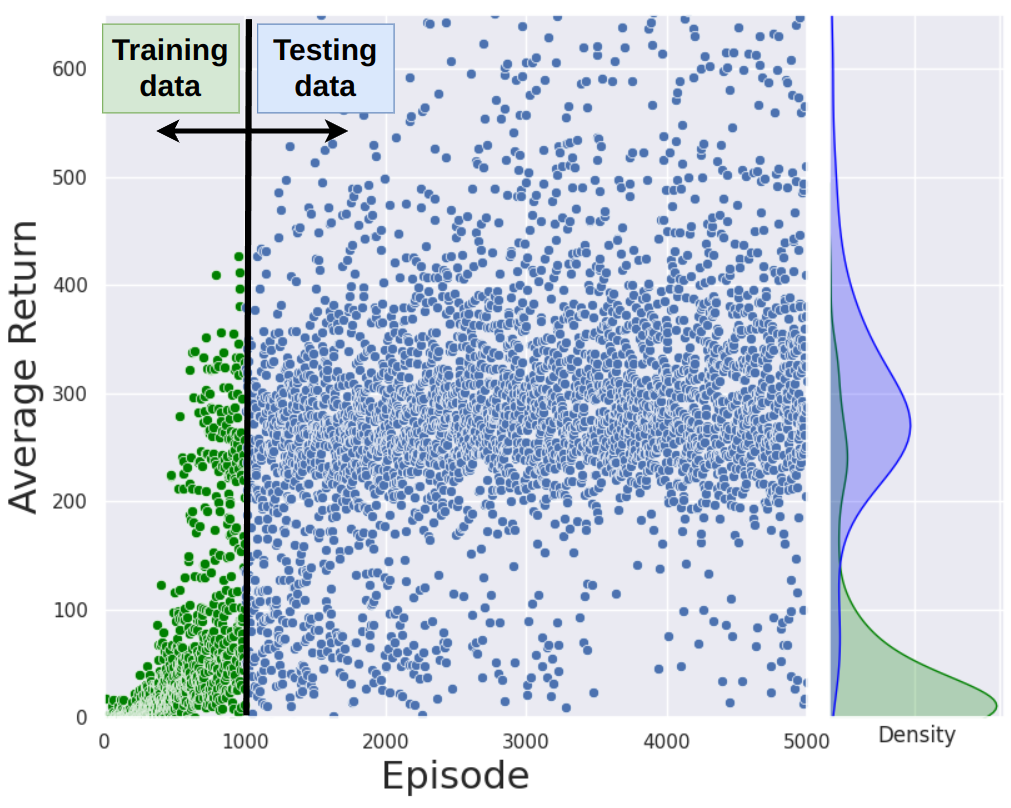}   &
        \includegraphics[width=0.225\linewidth]{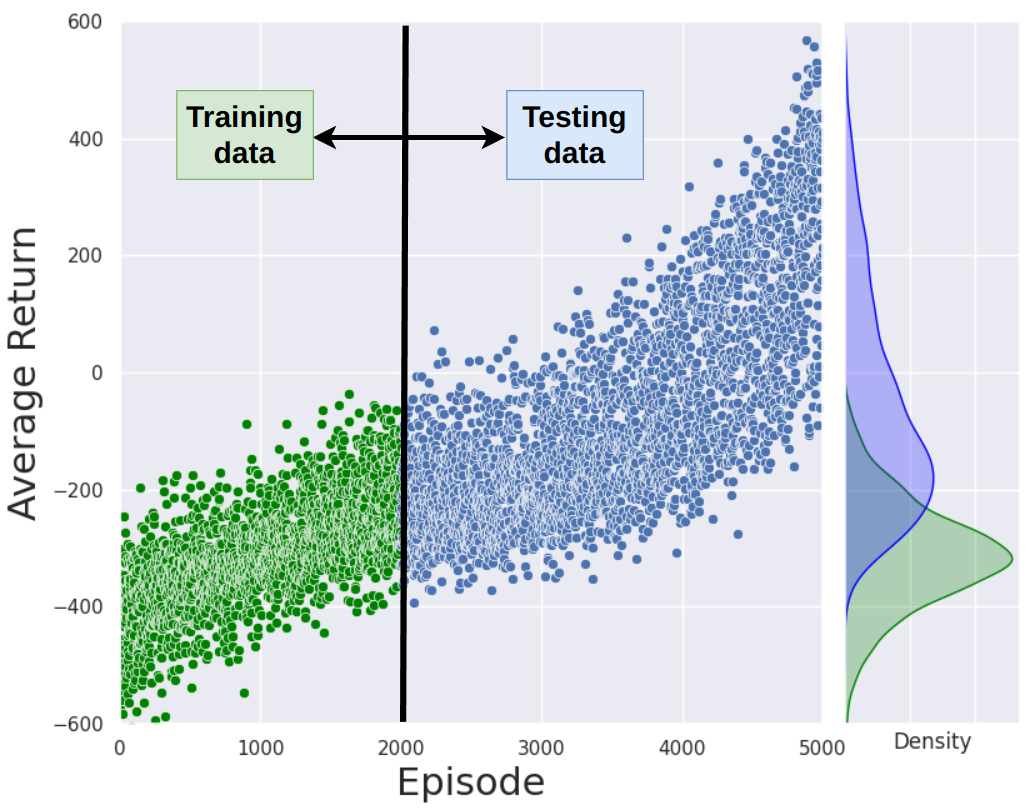}  \\
        
        (a) Inverted Pendulum & (b) Hopper & (c) Walker & (d) HalfCheetah    \\
    \end{tabular}
    
    \caption{This figure depicts our dataset's structure, highlighting the discrepancy between training data and the desired optimal distribution sought by the policy for improved rewards. Specifically, this setup is designed for evaluating OOD adaptation using low-quality data. Here, MDP tuples are sourced from green distributions (e.g., low-quality data) for training and blue distributions for testing.
    }
    \label{fig:data_composition}
\end{figure*}

\subsection{Causal Predictive Power of MOOD-CRL} \label{section:experiments_A}
We demonstrate the stability and predictive performance of MOOD-CRL in OOD scenarios using a straightforward discrete environment: FrozenLake \cite{1606.01540}. We compare the predictive outcomes with those generated by a standard MLP-based prediction by a non-extensive quick demonstration to reveal the fundamental insight of MOOD-CRL, as the comprehensive empirical studies are saved in its application in offline RL, the focus of this paper. We refer the reader to \cref{fig:errorplot} in \cref{Ap:errorplot} for the continuous environment.

\subsubsection{Experimental Design}
We commence with $(15 \times 15)$ deterministic FrozenLake domain as depicted in \cref{fig:frozenlake}. In this environment, the agent initiates at the top-left corner (state = 0) to reach the goal, bottom-right, state (state = 224). The agent has four available actions: moving left, down, right, and up. Notably, any actions directed towards obstacles or the boundary of the domain result in remaining in the same state. To evaluate the effectiveness of our model regarding causal inference, we constructed the training dataset by selecting the bottom 80\% of MDP tuples: state-action pairs, $(s,a)$, where $s \geq 45$ and $a \subset \mathcal{A}$; other remaining part is test dataset. Our primary focus lies in understanding whether the model can capture the fundamental dynamics of this domain, i.e., moving upwards entails subtracting 15 from the previous state or identifying the boundaries on the left and right sides.

\subsubsection{Evaluation}
We measured the model's error using the L1 norm for the transition dynamics of state in comparison to the standard MLP-based predictions. As depicted in Figure \ref{fig:Discrete}, causal normalizing flow effectively captures the dynamics, showcasing a notably reduced error attributed to its causal prowess. The substantial difference in error primarily stems from the susceptibility of the MLP when confronted with unseen inputs, $(s,a)$ where $s < 45$, leading to erroneous predictions. While causal normalizing flow yields significantly reduced error rates in detecting boundaries and obstacles, it is important to acknowledge that the upper boundary and obstacles are inherently not discoverable due to the absence of information in the dataset. However, the left and right boundaries hold the potential for inference. This proves the potential of causal normalizing flow as a world model effectively capturing the dynamics with fairly accurate outputs to unseen inputs.


\begin{figure*}[h!]
    \centering
    \begin{tabular}{c c c}
        \multicolumn{3}{c}{\includegraphics[width=0.9\linewidth]{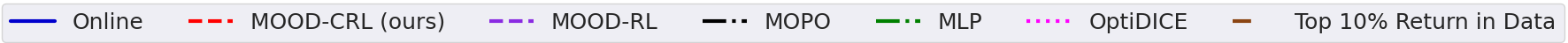}}\\
        \hline\hline
        \multicolumn{3}{c}{InvertedPendulum}\\
        \includegraphics[width=0.31\linewidth]{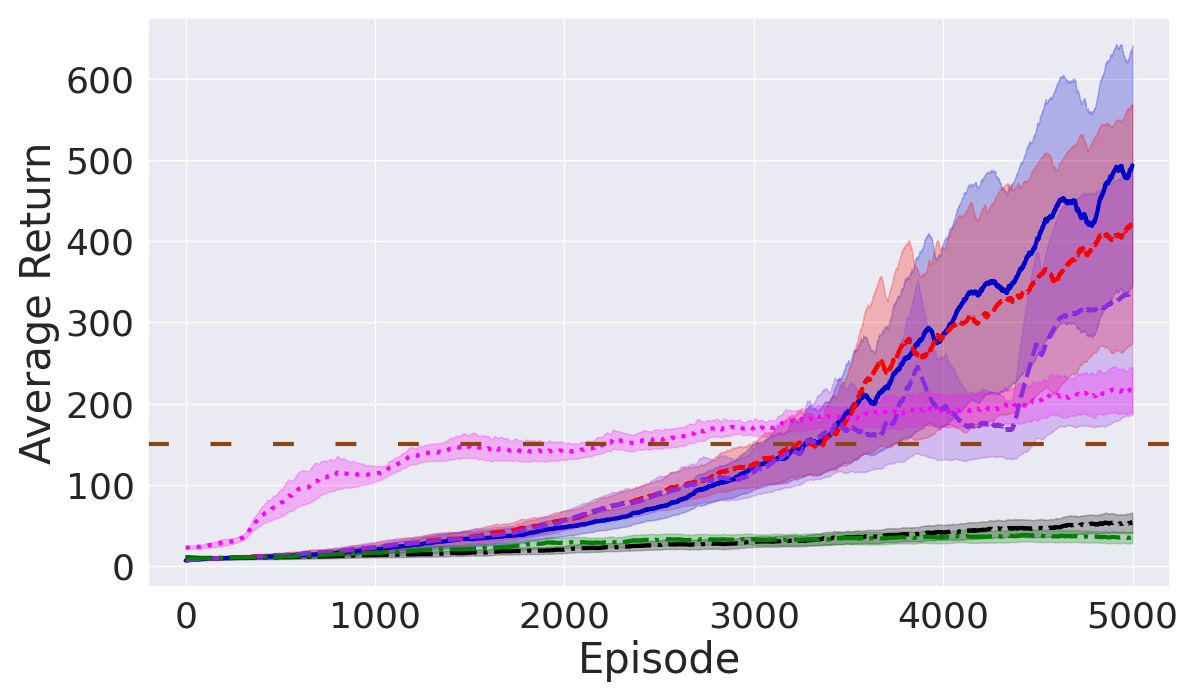}    &
        \includegraphics[width=0.31\linewidth]{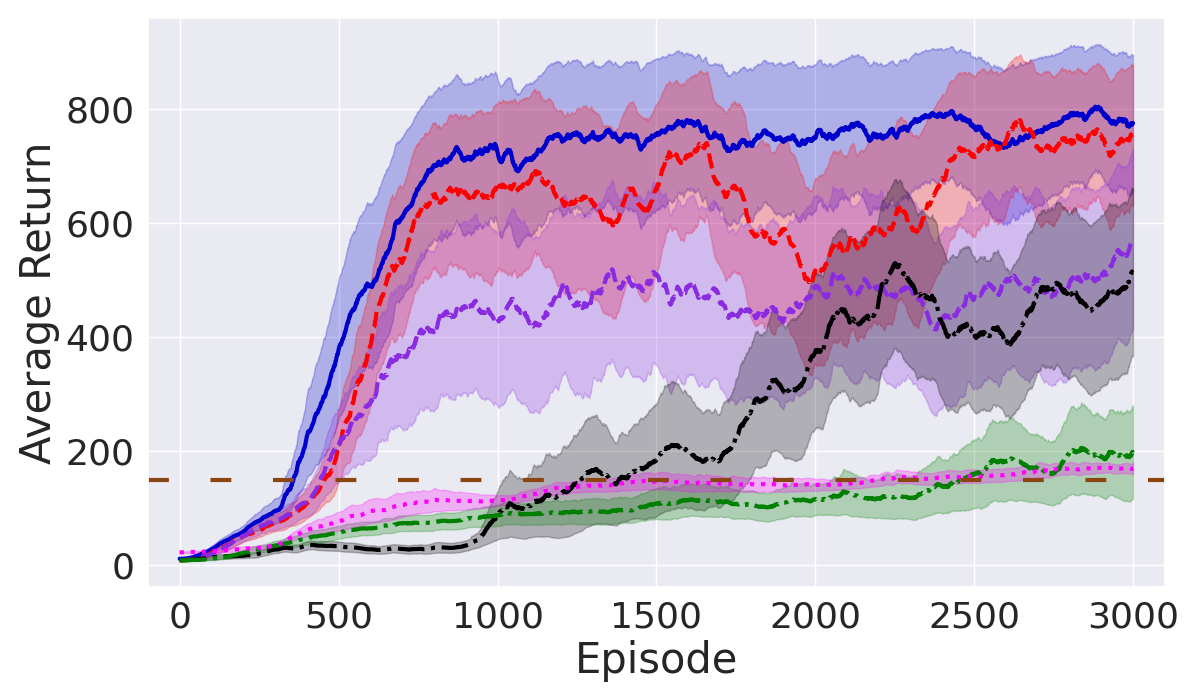}    &
        \includegraphics[width=0.31\linewidth]{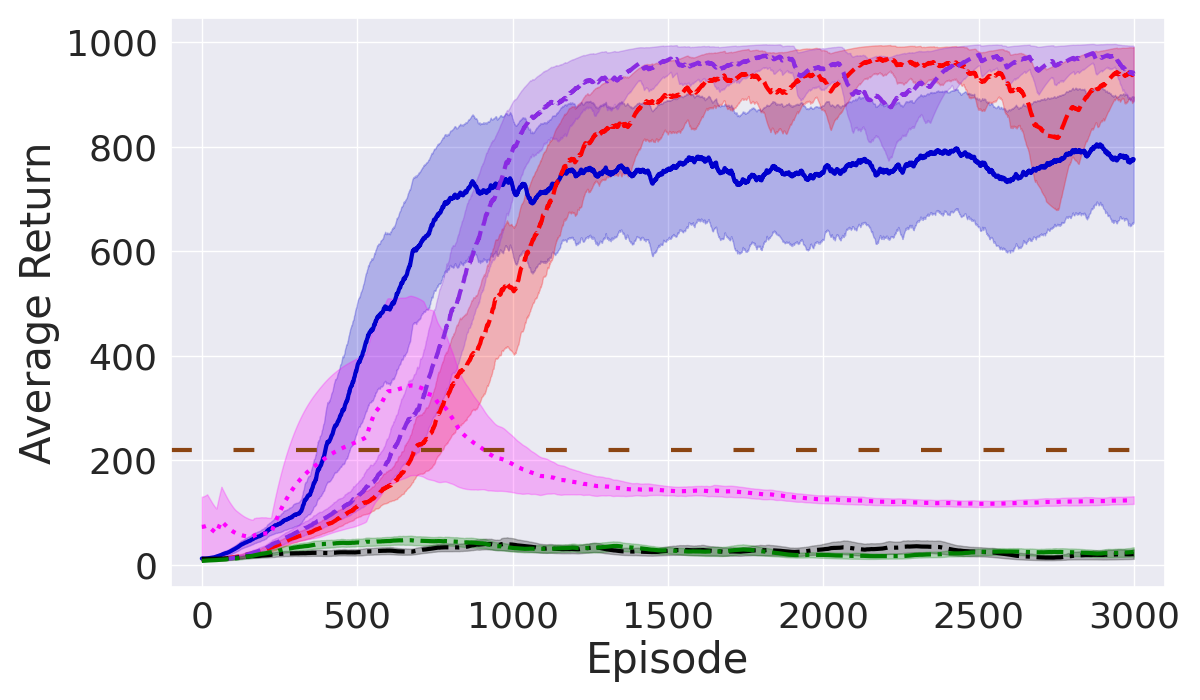}    \\
        
        \hline\hline
        \multicolumn{3}{c}{Hopper}\\
        \includegraphics[width=0.31\linewidth]{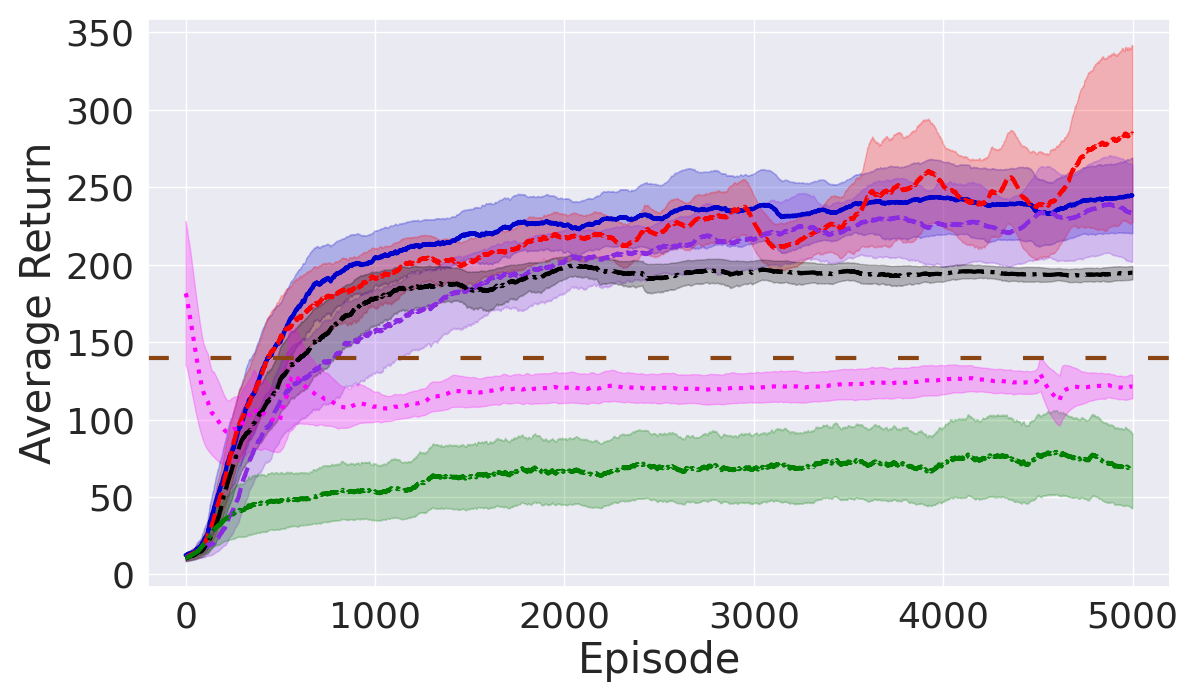}    &
        \includegraphics[width=0.31\linewidth]{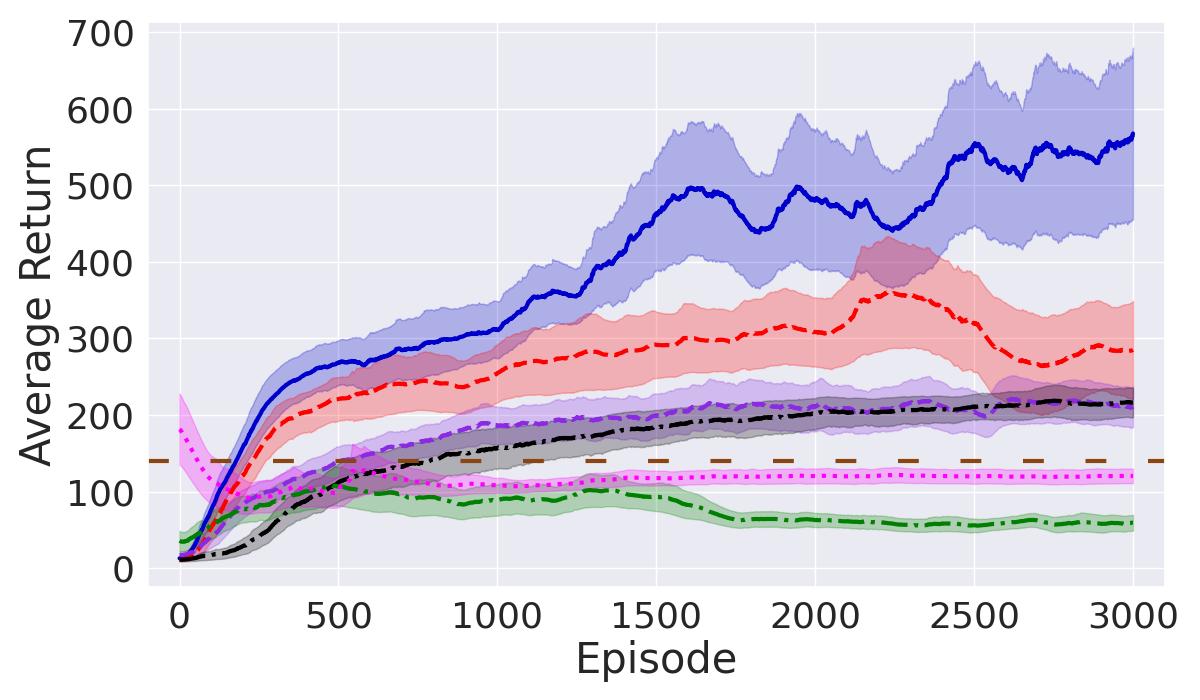}   &
        \includegraphics[width=0.31\linewidth]{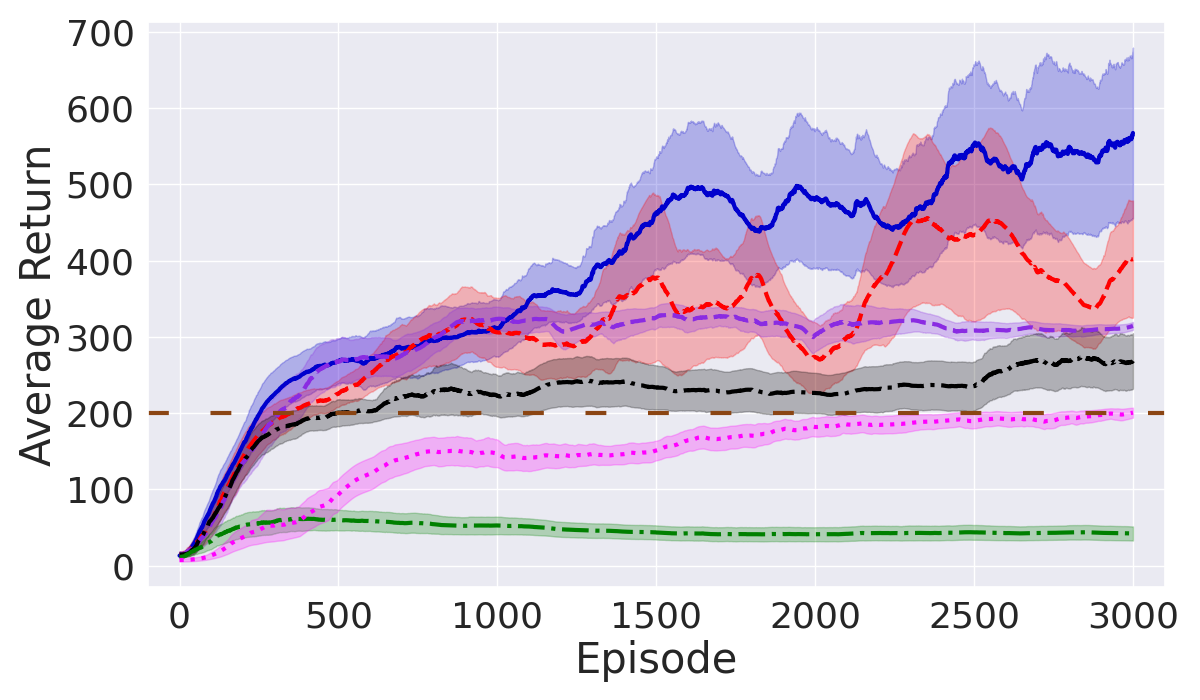}   \\

        \hline\hline
        \multicolumn{3}{c}{Walker}\\
        \includegraphics[width=0.31\linewidth]{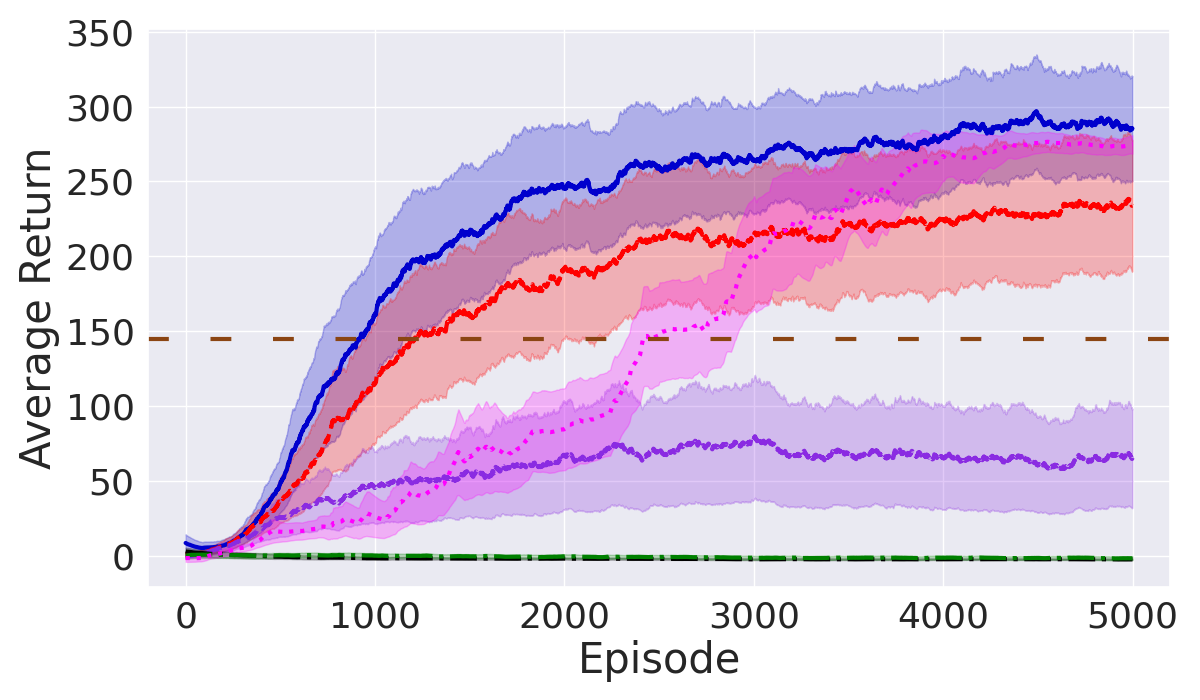}    &
        \includegraphics[width=0.31\linewidth]{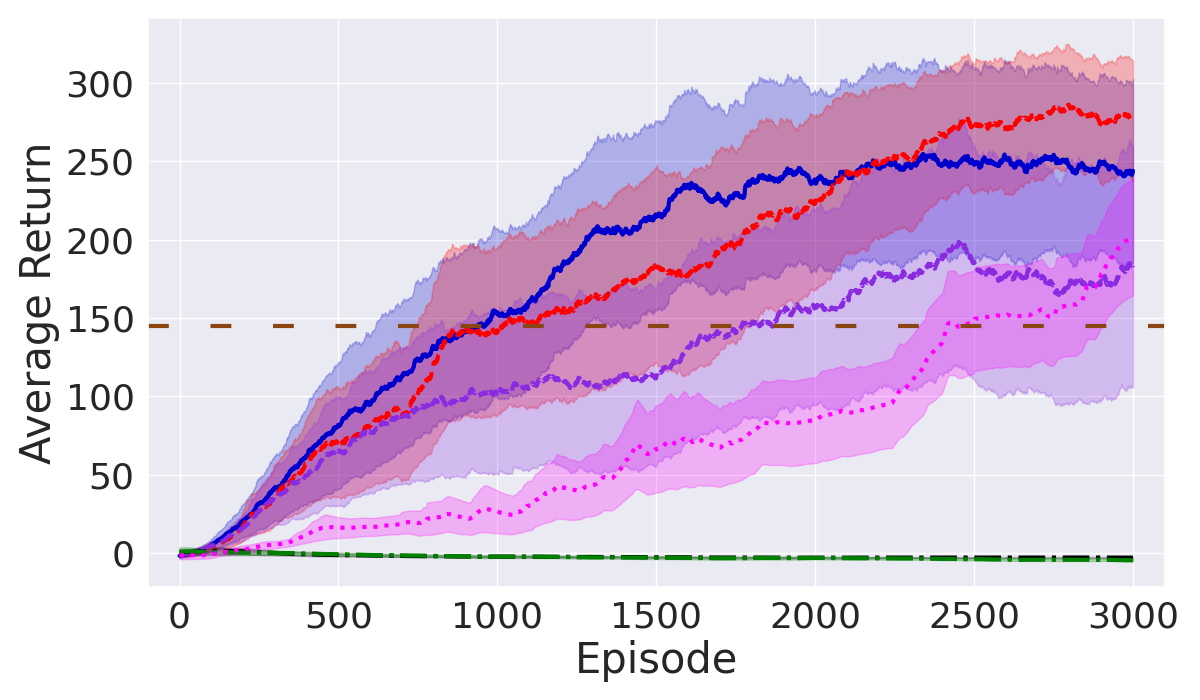}   &
        \includegraphics[width=0.31\linewidth]{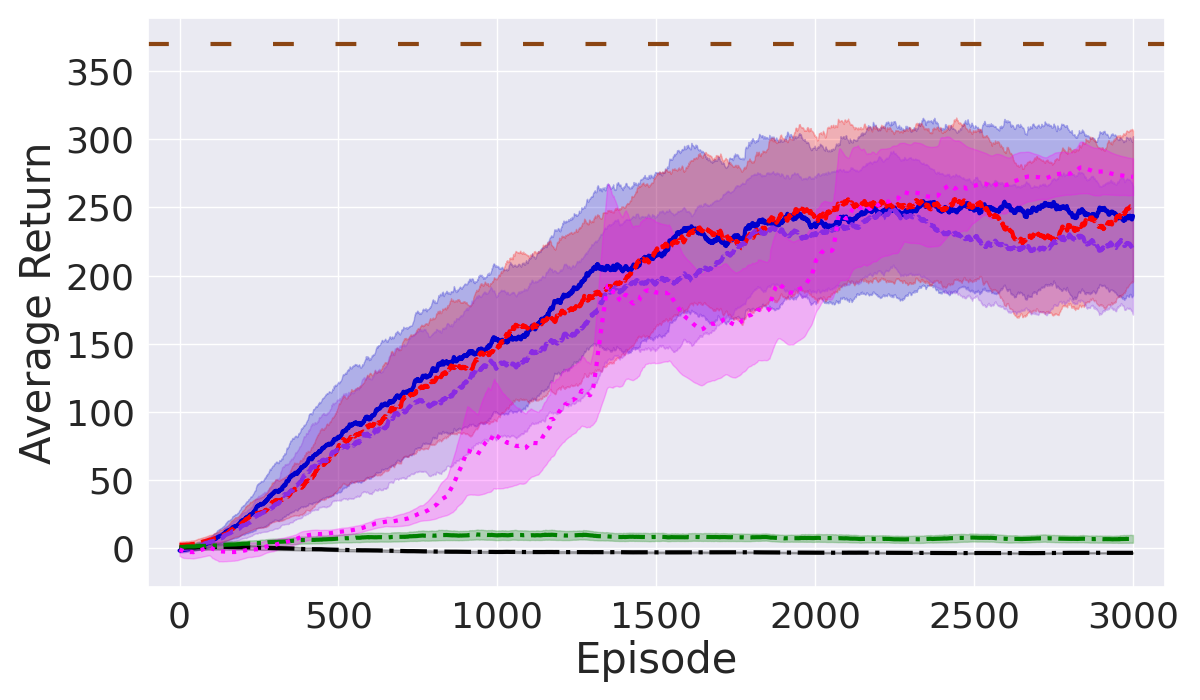}   \\

        \hline\hline
        \multicolumn{3}{c}{HalfCheetah}\\
        \includegraphics[width=0.31\linewidth]{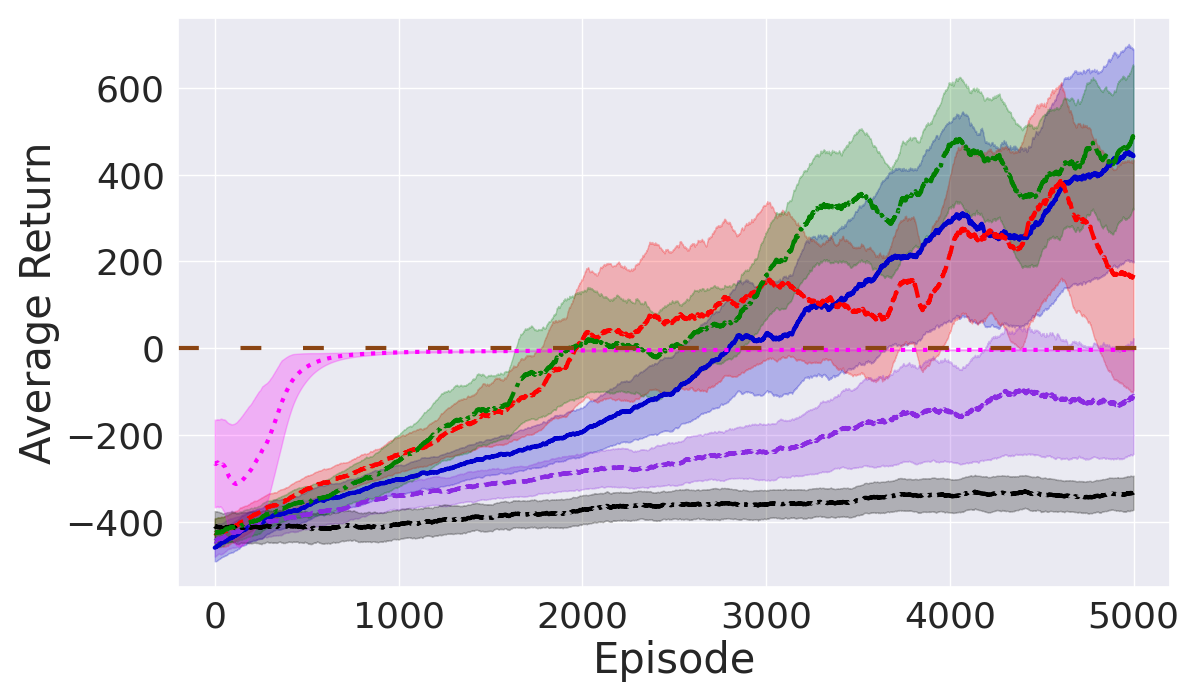}   &
        \includegraphics[width=0.31\linewidth]{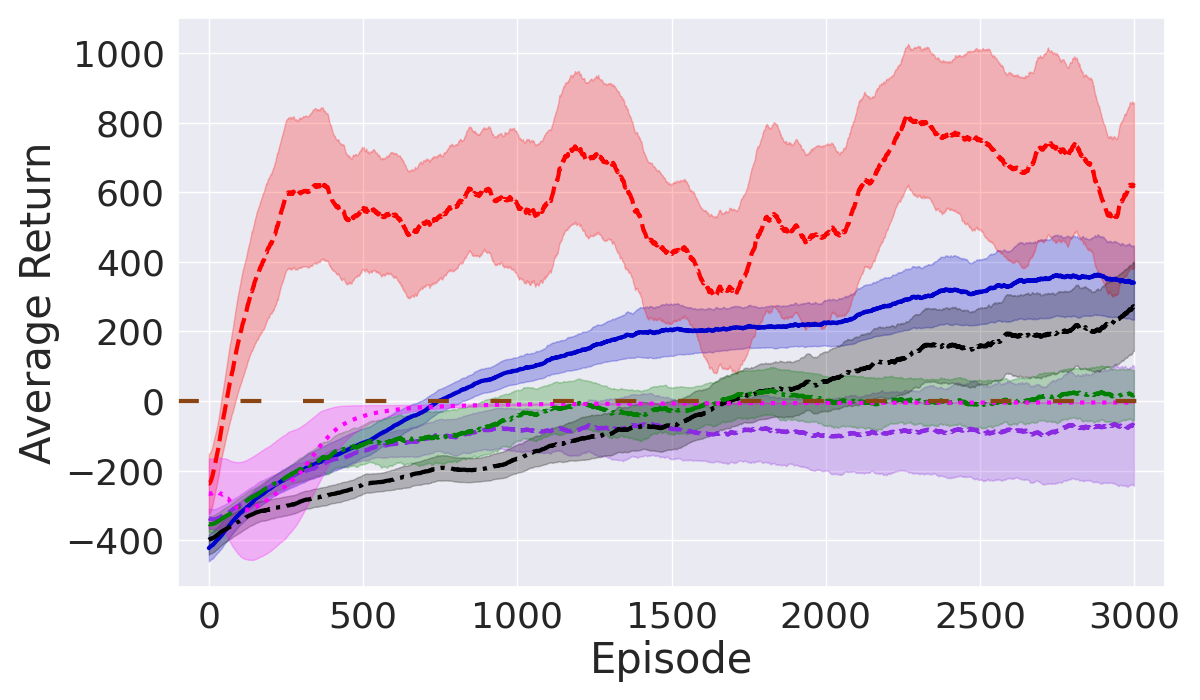}    &
        \includegraphics[width=0.31\linewidth]{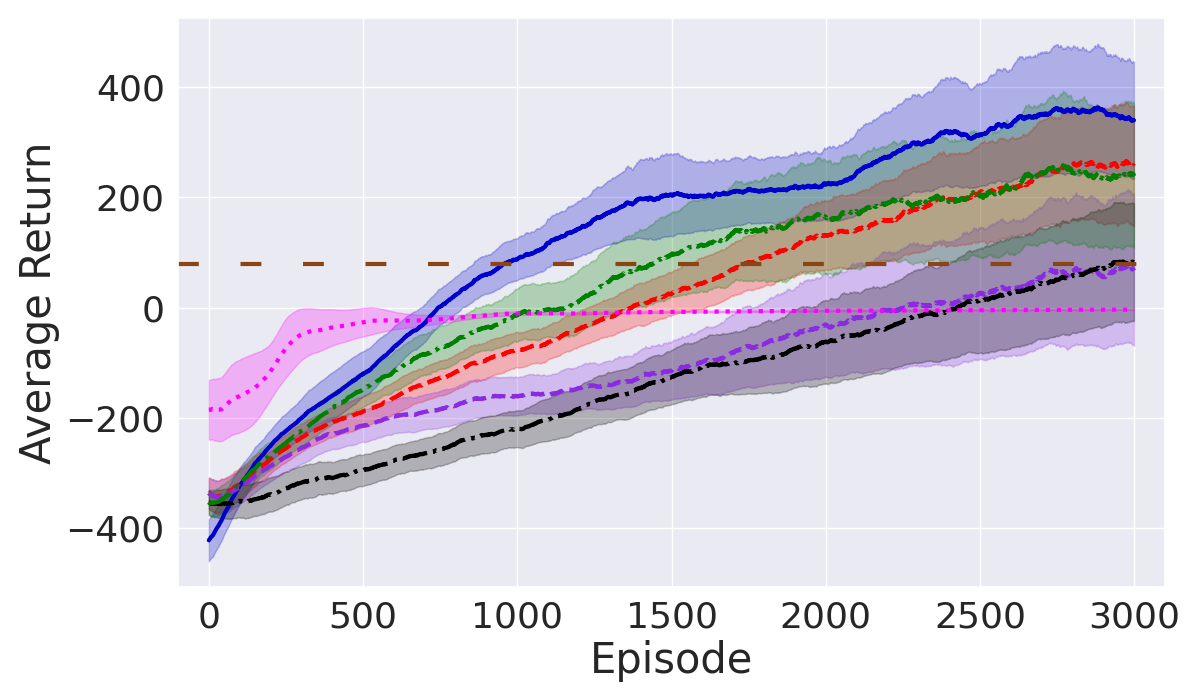}   \\
        
        (a) REINFORCE (low-quality data)  & (b) PPO (low-quality data) & (c) PPO (medium-quality data) \\
    \end{tabular}
    
    \caption{The figure depicts the average episodic returns (higher is better) over 5 seeds with standard errors as a shadow. Each row represents one domain while each column represents one pair of the base RL algorithm--data quality combination. The horizontal solid line represents the optimal return (\textbf{top 10\%}) found in the training data. The online learning curve (blue) is also plotted here for benchmarking. Our method, MOOD-CRL (shown in red), closely mirrors the online learning curve or its lower bound for all environments. MOPO (black) avoids OOD via uncertainty quantification, but it struggles to learn optimal policies in all environments and even fails in InvertedPendulum and Walker due to strong OOD constraints which are removed from our approach. Furthermore, except for Walker, the model-free OptiDICE exhibits high sensitivity to the provided dataset, which is unfavorable. With MOOD-CRL surpassing other methods, its non-causal counterpart (i.e., NF) also demonstrates competitive results across all domains, highlighting the structural superiority of our novel approach for OOD adaptation: base-distributional learning with a decoding network. Overall, our approach and its variants can consistently achieve high performance over baselines across different domains and settings. Conversely, MLP-based methods such as MOPO and neural networks reveal weaknesses in OOD predictions, which prevent learning high-performing policies.
    }
    \label{fig:whole_data}
\end{figure*}

\subsection{Extensive Assessments in Robotic Control Tasks}
In this section, our primary focus is on how the learned causal model can improve the performance of offline RL by predicting the transition dynamics and reward in the high-reward state-action space in \cref{fig:whole_data}. We offer thorough insights into how the architecture of MOOD-CRL improves offline learning performance over others. As it's crucial to understand how model errors and predictive dynamics can be managed within existing RL algorithms to ensure satisfactory performance, we evaluate our baselines using two different on-policy algorithms: the basic policy gradient method of REINFORCE \cite{sutton1999policy} and the state of the art policy gradient method of Proximal Policy Optimization (PPO) \cite{schulman2017proximal}. Furthermore, we empirically determined the parameter $c$ in \cref{eqn:truncate} to facilitate an OOD adaptation without degrading the policy to mitigate erroneous predictions. 


\subsubsection{Experimental Design}
To assess the OOD adaptation, we generate training data consisting of tuples $(s,a,s',r)$ explored by REINFORCE. Specifically, the training dataset was curated to include state-action pairs observed within the first 2000 episodes, as illustrated by the return distribution in \cref{fig:data_composition}. We refer to this as a low-quality dataset. While medium-quality data was collected from episodes 2000 to 3000, its size is maintained as the same as the low-quality dataset. This is for the ablation study and is explored in the next section.

This way of composing data facilitates a gradual learning process, akin to how infants learn to crawl before they walk. Afterward, the evaluation of OOD adaptation depends on the agent's ability to recognize improved reward zones once it has gained a general understanding of the given tasks. {The details of data composition are depicted in \cref{fig:data_composition}, while its t-SNE visualization \cite{van2008visualizing} is provided in \cref{fig:tsne_data} in the Appendix, offering a sound justification for this choice.}

\subsubsection{Evaluation}
As shown in \cref{fig:whole_data} and for low-quality data (first two rows), MOPO performs poorly when combined with REINFORCE. While it achieves competitive results with the PPO algorithm, it falls short of matching the performance of MOOD-CRL in every task. Additionally, in HalfCheetah, MOPO performs worse than traditional MLP. This is attributed to MOPO's tendency to underestimate OOD scenarios, resulting in penalized rewards that hinder policy improvement. This suggests that MOPO's sensitivity to policy learning can lead to inconsistent training results. Conversely, model-free OptiDICE generally does not achieve the same level of performance as model-based methods. This suggests that model-based approaches offer a more principled solution than model-free gradient estimation.

In contrast, our method, MOOD-CRL, closely tracks the online learning curves or at least their lower bounds. This holds across both REINFORCE and PPO. Additionally, MOOD-RL demonstrates highly competitive performance compared to MOPO in several domains, highlighting the benefits of base-distributional learning with interpretative tools such as normalizing flow. However, we observed a spike in our method for achieving returns exceeding the online learning curve (e.g., especially in HalfCheetah, PPO (low-quality data)), which is unexpected. We attribute this to slight errors in state prediction and corresponding policy actions that resonated to be high-quality state-action pairs leading the policy to converge rapidly to the optimal policy. As this appears to be an isolated case and no performance degradation is observed, we conclude that MOOD-CRL's predictions remain reasonably accurate for the entire domain.

\subsection{Ablation on Different Data Quality} \label{section:exp_3}
To assess the sensitivity of each baseline and ours to varying data quality, we conduct an ablation study by including medium-quality data (last row of \cref{fig:whole_data}) in addition to its low-quality counterpart. This differs from the previous method of employing low-quality data inputs, which involves utilizing knowledge from early iterations and assessing its OOD adaptation. Instead, we now task the algorithm with medium-quality data to learn a high-performing policy.


\subsubsection{Evaluation}
In the cases of Inverted Pendulum, Walker, and HalfCheetah, it is evident that MOPO and MLP fail to exhibit even minor improvements. This suggests that the predictions made by these models do not facilitate any OOD adaptation for optimal rewards. Conversely, ours, including MOOD-RL, demonstrate stable and consistent learning curves. This solves the previous inferiority of the model-based approach against model-free where given data lacks wide coverage as well as underscores the advantage over other model-based approaches.

\section{Related Works} \label{sec:related_work}
Offline reinforcement learning has been attracting much attention and effort in the community, with several notable methods emerging for its resolution. This includes both model-based and model-free algorithms, with resolutions on overcoming the distributional shift via constraints to match the training data distributions. It has been an active search to explore more sophisticated methods for incorporating such constraints via 1) policy constraint, 2) importance sampling, and 3) uncertainty estimation. Thus, we summarize the previous trend below with the taxonomy proposed by \cite{prudencio2023survey}.

\subsection{Model-free algorithms}
\subsubsection{Policy Constraints}
Works in \cite{fujimoto2019off, kumar2020conservative, wu2019behavior} introduced additional constraints to align the learning policy with the behavior policy, requiring the parameters of the behavior policy. These methods then employ a \emph{f}-divergence metric to quantify distributional divergence between the behavior policy and the learning policy to implicitly match the two distributions. Specifically, \cite{fujimoto2019off} estimates behavior policy via supervised regression, while \cite{kumar2020conservative} suggested support matching over distribution matching to prevent OOD actions, which was proved to be effective by exploiting only good actions in the data.

\subsubsection{Importance Sampling}
Additionally, another line of research seeks to estimate the distributional discrepancy between the behavior policy and the learning policy \cite{lee2021optidice, lee2022coptidice}. This discrepancy estimation aids in accurately evaluating the learning policy's performance, facilitating the identification of optimal rewards by aligning the expectation under the state marginal of the dataset with that of the learning policy. This approach is complemented by reward regularizers to penalize OOD occurrences as well.

\subsubsection{Uncertainty Estimation}
Alternatively, certain approaches directly estimate the uncertainty linked to the policy's actions given a state, aiming to alleviate OOD constraints and mitigate excessive conservatism. The estimated uncertainty can substitute the previous reward regularizer in policy constraint methods, specifically targeting high uncertainty in OOD regions, while allowing exploration in OOD regions with low uncertainty. {Our work fits within this taxonomy through robust uncertainty quantification achieved by normalizing flow.}

\subsection{Model-based algorithms}
Like our approach, model-based offline RL centers on estimating transition dynamics and reward functions. For instance, \cite{yu2020mopo} employs a supervised learning approach with explicit uncertainty quantification in transition dynamics to underestimate rewards and avoid OOD issues deliberately. These models, including recent advances of Transformer \cite{vaswani2017attention}, are commonly utilized as trajectory planners \cite{janner2021offline}.

\subsection{Others}
There are multiple other studies to enhance offline RL, such as  Lyapunov stability and control-invariant sets~\cite{kang2022lyapunov}, invariant representation learning~\cite{qi2022data}, mutual information regularizer~\cite{ma2022mutual}, anti-exploration~\cite{rezaeifar2022offline} to penalize OOD states/actions. However, the degree of conservatism to avoid overestimation is still ambiguous and can be sub-optimal. A straightforward extension is proposed in~\cite{hong2022confidence} by sampling and learning under multiple conservatism degrees. Another notable issue in offline RL is how to learn policies from data generated by multiple policies (with varying sampling distributions and performance) for the task~\cite{Kumar}. 

\section{Conclusions and Discussions} \label{section:conclusions}
While earlier advancements have centered on OOD constraints in the offline RL framework, our approach introduces an OOD-adapting algorithm empowered by causal inference and accurate detection of erroneous predictions in OOD space. This is accomplished by eliminating prior OOD penalization to address distributional shifts, while also preventing policy degradation from erroneous predictions. Our approach introduces a novel model architecture for offline RL, merging causal normalizing flow with a conventional MLP. This innovative formulation enables the use of normalizing flow to address offline RL challenges in a model-based approach, which has not been previously explored due to its structural constraints (bijective nature) for dynamic modeling, albeit its appealing properties. Normalizing flow is leveraged for the benefits of causal inference, OOD detection, and the ability to handle multi-modal distributions. Notably, our experiments demonstrate that the learning curve of our model closely resembles that of online training or its lower bound for all testing environments. This research offers a pioneering approach to model-based offline RL for out-of-distribution adaptation, emphasizing the importance of base-distributional learning with interpretative tools such as normalizing flow and VAE \cite{kingma2013auto}. {This property is particularly appealing as it addresses the existing weakness of model-based learning compared to the model-free approach. In particular, given a narrow coverage of data, the model-free approach was favored due to its ability to closely mimic the distribution of given data, while the model-based approach, as demonstrated in \cref{section:exp_3}, often performed poorly. This was because the given data might not provide sufficient initial knowledge to initiate learning of the primal steps required to achieve higher returns. However, our approach demonstrates OOD-adapting performance, which can be beneficial even in scenarios with limited datasets where model-free algorithms have traditionally excelled.} Furthermore, we anticipate that our method is generalizable and applicable to OOD adaptation in various domains, as it combines generative and predictive models with causal inference capabilities.

While this paper introduces a novel research avenue for model-based offline RL, we acknowledge several limitations of our approach. These include the absence of theoretical support and scalability issues, with computational costs increasing as the dimensions of states and actions grow. The latter is particularly pronounced as passing the entire MDP tuple through the model becomes more susceptible to the curse of dimensionality. Additionally, we suggest several intriguing research avenues worth exploring further. a) It is promising to investigate further which base-distributional learning mechanisms, coupled with interpretative tools (normalizing flow or VAE), can enhance overall learning outcomes, such as Bayesian neural network \cite{neal2012bayesian}, Recurrent Neural Network (RNN) \cite{Rumelhart1986LearningIR}, or Transformer \cite{vaswani2017attention}. b) Moreover, constrained settings \cite{achiam2017constrained} even with environmental non-stationarity \cite{cho2024constrained} can be considered in offline RL under MOOD-CRL's umbrella. This is particularly crucial in safety-critical domains like autonomous driving and robotic manipulation, where safety breaches are intolerable, and non-stationary conditions may prevail. These requirements necessitate rigorous proof through constraint-satisfaction analysis in OOD cases, either with chance-constrained or worst-case safety criteria. {c) Lastly, one could bring other techniques to enhance offline RL by studying conformal predictions for uncertainty quantification calibration and physics-informed neuro-symbolic RL with data-invariant logic rules for extrapolation.}

\bibliographystyle{plain}
\bibliography{refs}

\appendix 

\begin{figure*}[t!]
    \centering
    \begin{tabular}{c c}
        \multicolumn{2}{c}{\includegraphics[width=0.20\linewidth]{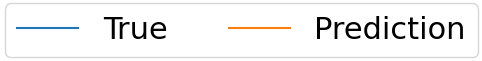}}\\
        \includegraphics[width=0.48\linewidth]{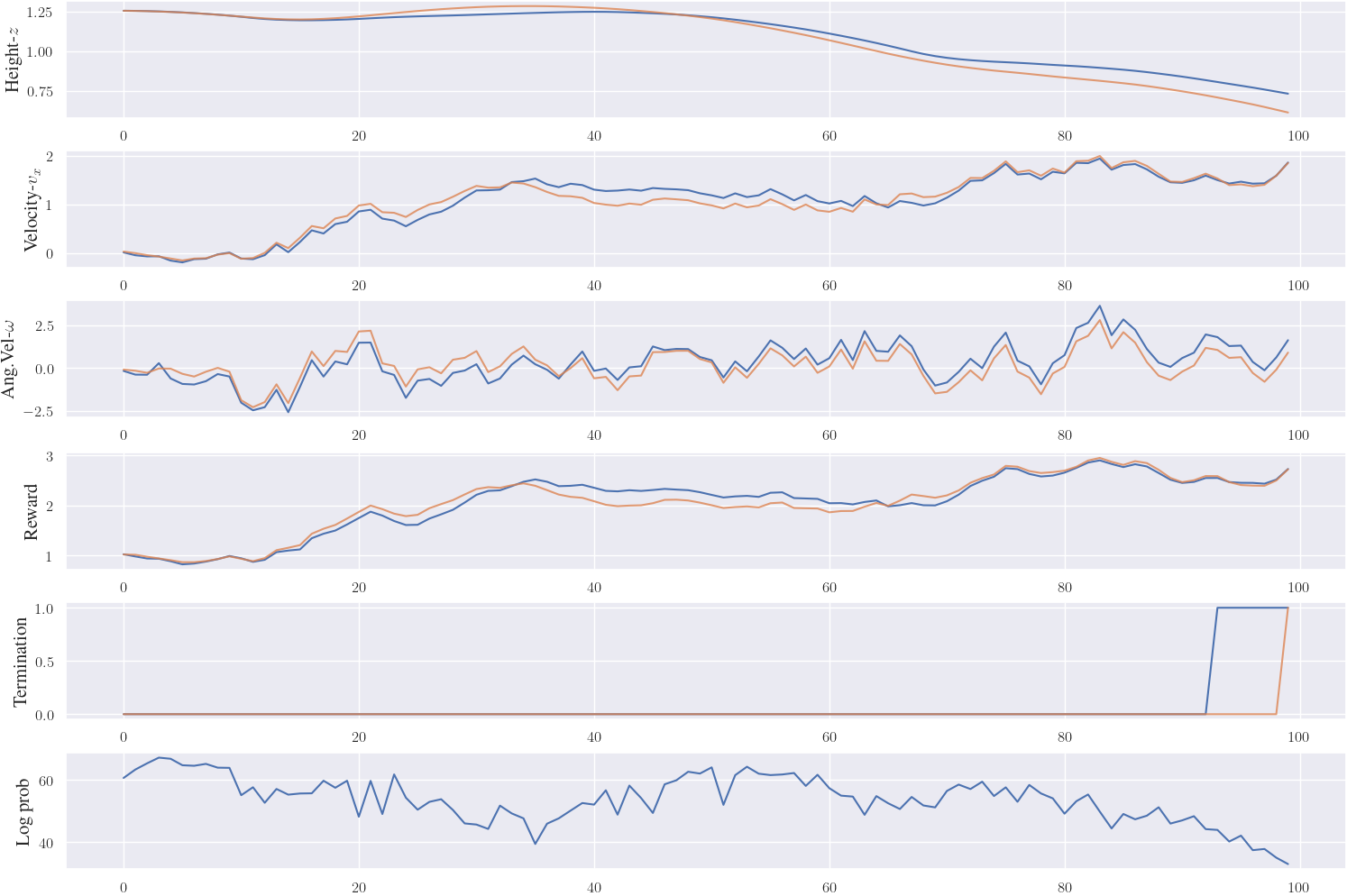}    &
        \includegraphics[width=0.48\linewidth]{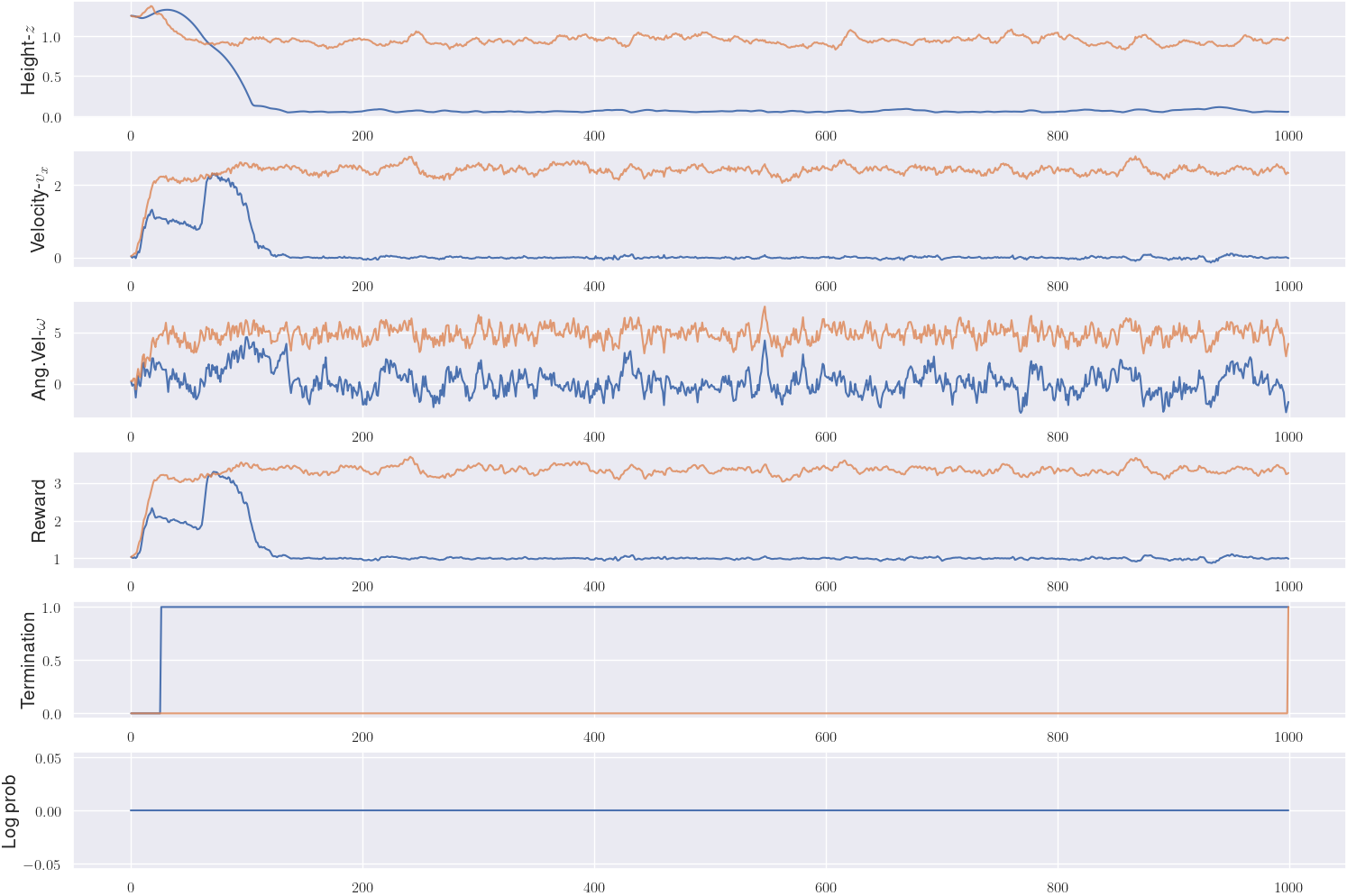}   \\
        
        (a) MOOD-CRL & (b) MLP-based predictions  \\
    \end{tabular}
    
    \caption{This illustrates the disparity between the true and predicted values in the Hopper. Key physical quantities of the torso of Hopper are plotted alongside rewards and termination (for those requiring termination prediction as well). Log probability is also included to validate the predictions made by our method. Notably, while MLP-based predictions may exhibit unknown patterns, our predictions form reasonable curves and patterns compared to the true values, albeit with some margins.}
    \label{fig:errorplot}
\end{figure*}

\section*{Experimental Details}
In this section, we outline the specific configurations of our experiments conducted in OpenAI Gym \cite{towers_gymnasium_2023}. We provide details on the composition of the training dataset and the underlying rationale. Additionally, we elaborate on the specific parameters employed for the baselines, including network size, key hyperparameters, and the causal graph used in our approach.

\begin{figure}[h!]
    \centering
    \begin{tabular}{c c}
        \multicolumn{2}{c}{\includegraphics[width=0.35\linewidth]{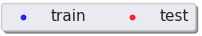}}\\
        \includegraphics[width=0.475\linewidth]{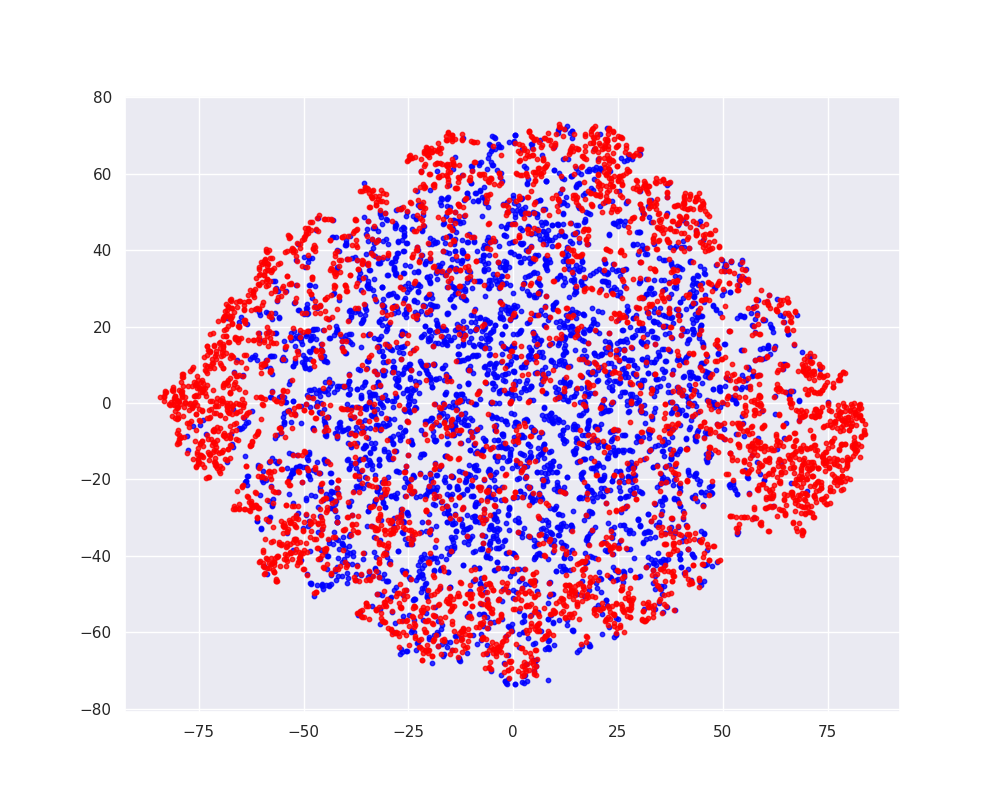}    &
        \includegraphics[width=0.48\linewidth]{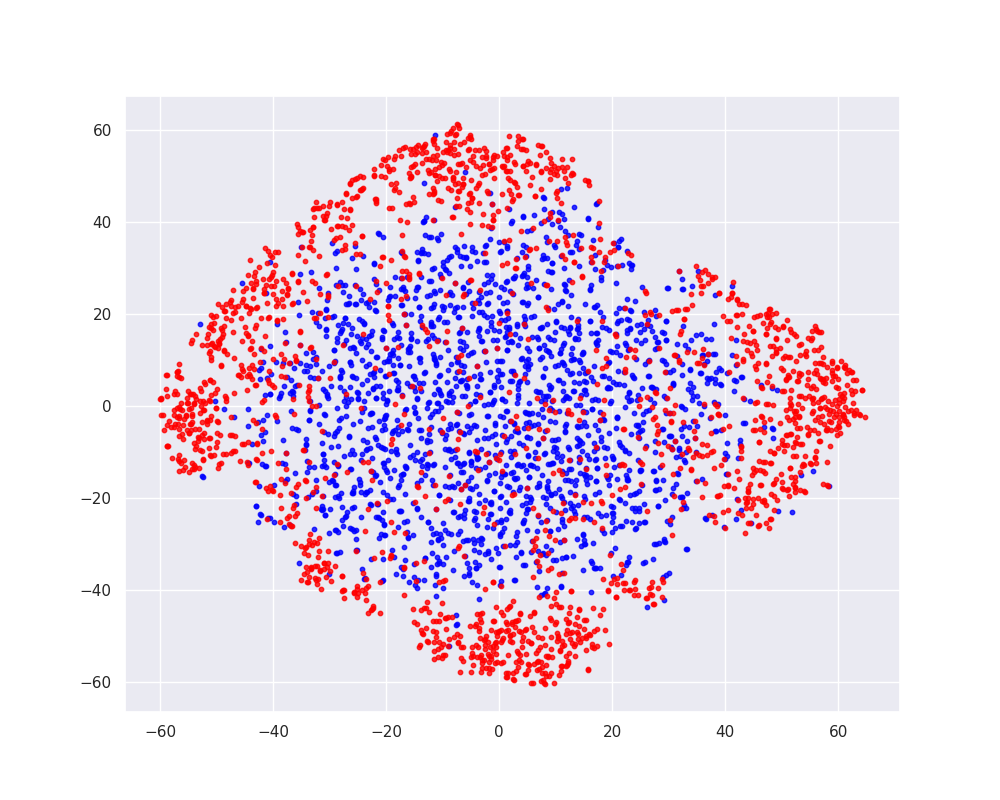}   \\
        
        (a) $(s,a,r)$  & (b) $(s,a,s')$  \\
    \end{tabular}
    
    \caption{The visualization showcases the t-SNE representation of HalfCheetah data, emphasizing the distributional shift between the training and testing datasets. The left figure captures the variance in reward dynamics, while the right figure visualizes the transitional shift in distribution.}
    \label{fig:tsne_data}
\end{figure}

\subsection*{State and Reward Predictions Curve} \label{Ap:errorplot}
For additional clarity, we provide a comparison of transition and reward dynamics between the trained model and the true oracle for the InvertedPendulum in \cref{fig:errorplot} for the sake of simplicity.

\subsection*{t-SNE Visualization of Dataset} \label{Ap:tsne_composition}

Using t-SNE visualization \cite{van2008visualizing}, the \cref{fig:tsne_data} validates our choice of data for OOD adaptation purposes.

\subsection*{Experimental Settings} \label{Ap:causal_graph}
Below, detailed hyperparameters for experiments and qualitative physics-informed causal graphs are discussed.
\subsubsection*{Hyperparameters}
\begin{itemize}
    \item \textbf{Optimizer:} Adam, Learning Rate: 1e-4
\end{itemize}

\begin{itemize}
    \item \textbf{CNF (Causal Normalizing Flow):}
        \begin{itemize}
            \item Flow model: Masked Autoregressive Flow \cite{papamakarios2017masked}
            \subitem Architecture: conditioner $c_i = $ (64, 64, 64) with 5 layers of transformation $F_1 \circ \cdots \circ F_5$
            \subitem Base Distribution: Normal
            \subitem Activation: elu
            \item Regularization: weight regularization
            \subitem Training Epochs: 2000
            \item Base model: MLP
            \subitem Architecture: neural network dimension of (512, 512, 512, 512)
            \subitem Activation: leakyrelu
            \item Regularization: weight regularization
            \subitem Training Epochs: 1000
        \end{itemize}
        
    \item \textbf{MOPO:}
        \begin{itemize}
            \item Architecture: Single ensemble with dimensions (512, 512, 512, 512)
            \item Activation: elu
            \item Regularization: weight regularization
            \item Training Epochs: 1000
        \end{itemize}
    
    \item \textbf{MLP:}
        \begin{itemize}
            \item Architecture: neural network dimension of (512, 512, 512, 512)
            \item Activation: leakyrelu
            \item Regularization: weight regularization
            \item Training Epochs: 1000
        \end{itemize}
    
    \item \textbf{Policy Network:}
        \begin{itemize}
            \item Architecture: neural network dimension of (64, 64)
            \item Algorithm: REINFORCE and PPO
                \subitem REINFORCE: lr = 1e-4
                \subitem PPO: K-epochs = 2, lr(policy) = 1e-4, lr(critic) = 3e-3, epsilon-clip = 0.2
        \end{itemize}

\end{itemize}

\end{document}